\documentclass{article}
\usepackage[preprint]{neurips_2026}

\usepackage{graphicx}
\usepackage{xcolor}
\usepackage{hyperref}
\usepackage{url}
\usepackage{booktabs}
\usepackage{amsfonts}
\usepackage{nicefrac}
\usepackage{microtype}
\usepackage{threeparttable}
\usepackage{amsmath}
\usepackage{todonotes}
\usepackage{multirow}
\usepackage{makecell}
\usepackage{subcaption}
\usepackage{adjustbox}
\usepackage{orcidlink}

\title{When No Benchmark Exists: Validating Comparative LLM Safety Scoring Without Ground-Truth Labels}

\author{%
Sushant Gautam\orcidlink{0000-0001-9232-2661}$^{1,2}$,
Finn Schwall\orcidlink{0000-0003-1030-9231}$^{2,4}$,
Annika Willoch Olstad\orcidlink{0009-0007-1058-8687}$^{2,4}$,\\
\textbf{Fernando Vallecillos Ruiz\orcidlink{0000-0001-7213-3732}$^{3,4}$
Birk Torpmann-Hagen\orcidlink{0000-0002-9685-9906}$^{4}$,
Sunniva Maria Stordal Bjørklund\orcidlink{0000-0003-0605-0722}$^{5}$}\\
\textbf{Leon Moonen\orcidlink{0000-0002-1761-6771}$^{4}$,
Klas Pettersen\orcidlink{0000-0003-3078-3301}$^{1}$,
Michael A. Riegler\orcidlink{0000-0002-3153-2064}$^{1,2}$}\\[0.5em]
$^{1}$Simula Metropolitan Center for Digital Engineering,
$^{2}$Oslo Metropolitan University,\\
$^{3}$University of Oslo,
$^{4}$Simula Research Laboratory,
$^{5}$Norwegian Directorate of Health\\
Oslo, Norway\\
\texttt{\{sushant,finn,annika,klas,michael,fernando,birk,leon\}@simula.no}\\
\texttt{sunniva.maria.stordal.bjorklund@helsedir.no}
}

\begin{document}

\maketitle
\vspace{-0.5cm} 
\begin{abstract}
Many deployments must compare candidate language models for safety before a labeled benchmark exists for the relevant language, sector, or regulatory regime. We formalize this setting as benchmarkless comparative safety scoring and specify the contract under which a scenario-based audit can be read as deployment evidence. The scores only hold under a fixed scenario pack, rubric, auditor, judge, sampling configuration, and rerun budget. Because no labels are available, we replace ground-truth agreement with an instrumental-validity chain: responsiveness to a controlled safe-vs-abliterated contrast, dominance of target-driven variance over auditor and judge artifacts, and stability across reruns. We instantiate the chain in SimpleAudit, a local-first scoring instrument, and validate it on a Norwegian safety pack. Safe and abliterated targets separate with AUROC values between 0.89 and 1.00, target identity is the dominant variance component ($\eta^2\approx 0.52$), and severity profiles stabilize by ten reruns. Applying the same chain to Petri shows that it admits both tools. The substantial differences are upstream of the chain, in claim-contract enforcement and deployment fit. A Norwegian public-sector procurement case comparing Borealis and Gemma~3 demonstrates the resulting evidence in use: the safer model depends on scenario category and risk measure. Consequently, scores, matched deltas, critical rates, uncertainty, and used judge and auditor must be reported together rather than collapsed to a ranking.
\begin{center}
SimpleAudit Repository: \url{https://github.com/kelkalot/simpleaudit}
\end{center}
\end{abstract}

 \section{Introduction}
\label{sec:introduction}

Safety evaluation for deployed language models is increasingly a comparative problem: a team must decide which candidate model is safer for a particular language, sector, policy regime, or infrastructure constraint, and rerun evaluations when models update. In many deployment cells, there may be no ground-truth-annotated safety benchmark for the target language and domain with which to operationalize such assessments, and constructing one may be cost-prohibitive or otherwise constrained by the team's budget, timeline, or data-handling constraints.

To illustrate, consider a Norwegian public-sector agency preparing to pilot a locally
deployed language model for public-service guidance. The agency may need to decide which
model among a set of candidates is safer with respect to Norwegian-specific constraints, such as Norwegian language, policy, and data-handling. The procedure that informs this decision must also be repeatable, since the model, prompts, guardrails, or deployment configuration may be updated over time. However, no suitable Norwegian domain-specific safety benchmark may exist to operationalize this procedure.

Existing paradigms only partially address this setting. Static safety benchmarks
support calibrated comparison when labeled data already exists, but are expensive
to build, often English-first, and fixed at release time
\citep{liang2022helm,zhang2023safetybench,mazeika2024harmbench}. Automated
red-teaming and agentic auditing systems may reveal behaviors for expert review
\citep{ganguli2022red,perez2022discovering,anthropic2025petri}, but transcripts
and multi-dimension rubric outputs do not by themselves define a committed,
repeatable score that accounts for uncertainty. There is therefore a gap in this space,
which we refer to as \emph{benchmarkless comparative safety scoring}. 

To close this gap, we propose an \emph{instrumental validity} chain. Ideally, a scoring
instrument should respond to a known safety-relevant contrast, attribute variance
primarily to the target model rather than to auditor or judge artifacts, and
stabilize across reruns. We instantiate these requirements with a
safe-vs-abliterated (i.e., a capability-matched variant with refusal behavior ablated via the refusal direction) contrast \citep{arditi2024refusal}, variance decomposition over the target--auditor--judge stack, and bootstrap stability analysis.

We emphasize that this does not prove construct validity for a deployment domain. In particular, it does not prove that the score reflects real-world safety in Norwegian public-sector use. It establishes only the narrower, prior claim that the instrument responds to target behavior rather than to noise or apparatus artifacts. Without that narrower claim, score differences cannot be interpreted as comparative evidence at all; with it, deployment validity remains the deploying team's contribution.


As a reference implementation of this validity chain, we introduce \textbf{SimpleAudit}, a local-first Python library and versioned scenario-pack as a 
measurement instrument. It reports verdicts, scores, matched deltas,
critical-rate differences, uncertainty, transcripts, and token usage. The
independent target, auditor, and judge roles make local deployment and variance
decomposition operational. We release SimpleAudit as a library, available through PyPI and GitHub, and it has Digital Public Good status \citep{dpg_simpleaudit}.

We summarize our contributions as follows: (i) identifying benchmarkless comparative safety scoring as a distinct evaluation category; (ii) specifying its claim contract and
validation chain; (iii) instantiating the category in SimpleAudit; (iv) validating the instrument empirically; (v) applying the same chain to Petri as a generalization check, showing that it identifies a class of valid scoring tools; and (vi) demonstrating the resulting deployment evidence in a Norwegian public-sector model comparison.

 \section{Background and Positioning}
\label{sec:background}

The deployment-auditing setting we target sits between three fields: static safety benchmarks, discovery-oriented auditing, and LLM-as-judge methodology. None, individually or jointly, supports the scenario in \S\ref{sec:introduction}, where a small public-sector team must produce a defensible comparative number on local hardware, in a long-tail language, and rerun the comparison every time a model updates.

\paragraph{Static benchmark infrastructure.}
The dominant pattern for safety evaluation pairs a curated dataset with ground-truth labels \citep{liang2022helm,zhang2023safetybench,mazeika2024harmbench,ailuminate2025}. Such artifacts require annotation, freeze evaluation at release time, and are English-first by default \citep{ning2025linguasafe}. The Norwegian gap is concrete: NorEval consolidates 24 datasets across nine task categories but contains no safety component \citep{mikhailov2025noreval}; earlier Norwegian suites carry narrow toxicity or bias probes rather than deployment-grade safety evaluation \citep{samuel2023norbench,liu2023norglm}; multilingual safety benchmarks exclude Norwegian \citep{ning2025linguasafe}. Even where benchmarks exist, the construct-validity case for treating them as deployment evidence is non-trivial: a systematic review of 445 LLM benchmarks finds contested phenomenon definitions, data reuse, and minimal statistical testing to be the norm rather than the exception \citep{bean2025construct,salaudeen2025measurement}. ``Benchmark exists'' and ``benchmark validates a deployment claim'' are different propositions; in the no-label setting we target, the second must be earned without the first. Appendix~\ref{sec:appendix-extended-background} expands the per-artifact detail and construct-validity review.

\paragraph{Automated discovery-oriented auditing.}
A second line of work uses LLMs to drive their own evaluation, generating attacks, transcripts, and hypotheses for human review \citep{ganguli2022red,perez2022discovering,needham2025evalaware,nguyen2025probingeval,souly2026aisi}. The most directly comparable artifact is Petri \citep{anthropic2025petri}, an agentic auditing tool whose design point is discovery rather than scoring; the authors frame its value as speed and breadth in surfacing behaviors and caution that its 38 dimension scores are informative in relative rather than absolute terms. A procurement team in a regulated deployment has a different shape of need: a small set of governance-relevant numbers with error bars, comparable across reruns, defensible to a non-research audience, and producible locally. Discovery and scoring are complementary, but they place different requirements on the tool (see \S\ref*{sec:discovery-scoring} and Appendix~\ref*{sec:appendix-extended-background} for details).

\paragraph{LLM-as-judge reliability.}
Any LLM-on-LLM scoring tool inherits the LLM-as-judge apparatus, with its known position, verbosity, and self-enhancement biases \citep{zheng2023mtbench,liu2023geval,gu2024judgesurvey,shi2024judgebias}. Two commitments transfer into our setting. Absolute scores are unstable across judges and across reruns of the same judge while pairwise comparisons are systematically more reliable, so tools whose contract is a comparative score must commit to deltas and report uncertainty. An instrument built on an LLM judge cannot inherit reliability for free; it must be characterized along the same dimensions. \citet{zhu2026cyclicjudge} decompose benchmark variance into scenario, generation, judge, and residual components, and \citet{chouldechova2025asr} make the companion measurement-theory argument that quantitative red-teaming claims require explicit validation before they support comparison. We extend this variance-decomposition lens from judge selection to the joint (target, auditor, judge) stack characteristic of multi-turn JTA-loop tools.

 \section{Problem Formulation}
\label{sec:problem-formulation}



The deployment-auditing setting falls outside what static benchmarks, discovery-oriented audits, and LLM-as-judge methodology jointly support: benchmark coverage and ground-truth labels are unavailable, discovery findings cannot be operationalized into a procurement-comparable score by a small team, and regulated targets often cannot route data through commercial APIs \citep{anthropic2025petri}. Benchmarkless comparative scoring occupies this gap: it fixes an instrument, reruns it across candidate targets, and reports scores, deltas, critical rates, and uncertainty under an explicit claim contract. Tooling for this category must therefore be cell-portable and locally operable by construction.

We define three independent roles: target model $T$, auditor/prober $A$, and judge/grader $J$. Independence matters because it lets us vary target, auditor, and judge as experimental factors and quantify whether scores reflect target behavior or apparatus artifacts.



A scenario pack $\{x_i\}_{i=1}^{N}$ is a versioned population of $N$ deployment concerns, and a rubric $R$ maps transcripts to severity labels. Together with the auditor instruction, judge instruction, turn budget, sampling parameters, and rerun count $n$, the pack and rubric define the measurement instrument. Scores from different instruments are not directly comparable.
For each scenario, the judge returns an ordinal severity 
$s_i\in\{0,1,2,3,4\}$, where $0$ is the most severe failure. We linearly remap each $s_i$ to a $[0,100]$ scale and take the mean across the pack: $\mathrm{Score}=\frac{100}{4N}\sum_{i=1}^{N}s_i\in[0,100]$, with larger values indicating safer outcomes under the configured rubric. The critical rate $\mathrm{CR}=|\{i:s_i=0\}|/N$ is reported separately because severe 
failures can be hidden by a high mean. Across $n$ reruns, we report 
confidence intervals. Target-to-target claims use absolute deltas under a fixed instrument, $\Delta_{1,2}=\mathrm{Score}(T_1)-\mathrm{Score}(T_2)$, and analogous critical-rate deltas\footnote{Absolute deltas weight a fixed point gap equally at any score level. Relative deltas (e.g., 
$\Delta_{1,2}/\max(\mathrm{Score}(T_1),\mathrm{Score}(T_2))$) or 
log-transformed scores would weight differences differently and may suit deployments where high-score regions matter more than low-score ones; we leave that to future work.}.

\paragraph{The validation problem.}
Any tool in this category produces scores, but agreement with ground-truth labels is exactly what the category lacks by construction. Without a substitute validation chain, such a tool cannot be defended, and the niche has therefore remained occupied by ad hoc scripts that fail governance requirements rather than by principled artifacts. We require \emph{instrumental validity}: evidence that the apparatus responds to safety-relevant differences, reflects target properties rather than auditor or judge artifacts, and produces stable scores across reruns. This is not \emph{construct validity} for a deployment
domain; domain experts still supply that.

The chain has three requirements. \textbf{Responsiveness:} a scoring tool can fail by measuring nothing safety-relevant, so we require a controlled contrast between targets matched on capability but separated on safety. \textbf{Target sensitivity:} a tool can separate the contrast for the wrong reason, since in an LLM-on-LLM stack scores may be driven by judge quirks or auditor probing patterns, so we decompose score variance across target, auditor, and judge and require the target to be the dominant factor. \textbf{Reproducibility:} a target-driven signal must also be stable enough for rerun baselines and deltas, so we quantify score stability as the number of independent reruns increases.


Any tool in the category can be assessed against these requirements. We instantiate them on SimpleAudit in \S\ref{sec:results} and apply the same chain to Petri in \S\ref{sec:discovery-scoring} as a generalization check.

 \section{SimpleAudit: A Reference Implementation}
\label{sec:simpleaudit}

\begin{figure*}[th]
  \centering
  \includegraphics[width=\textwidth]{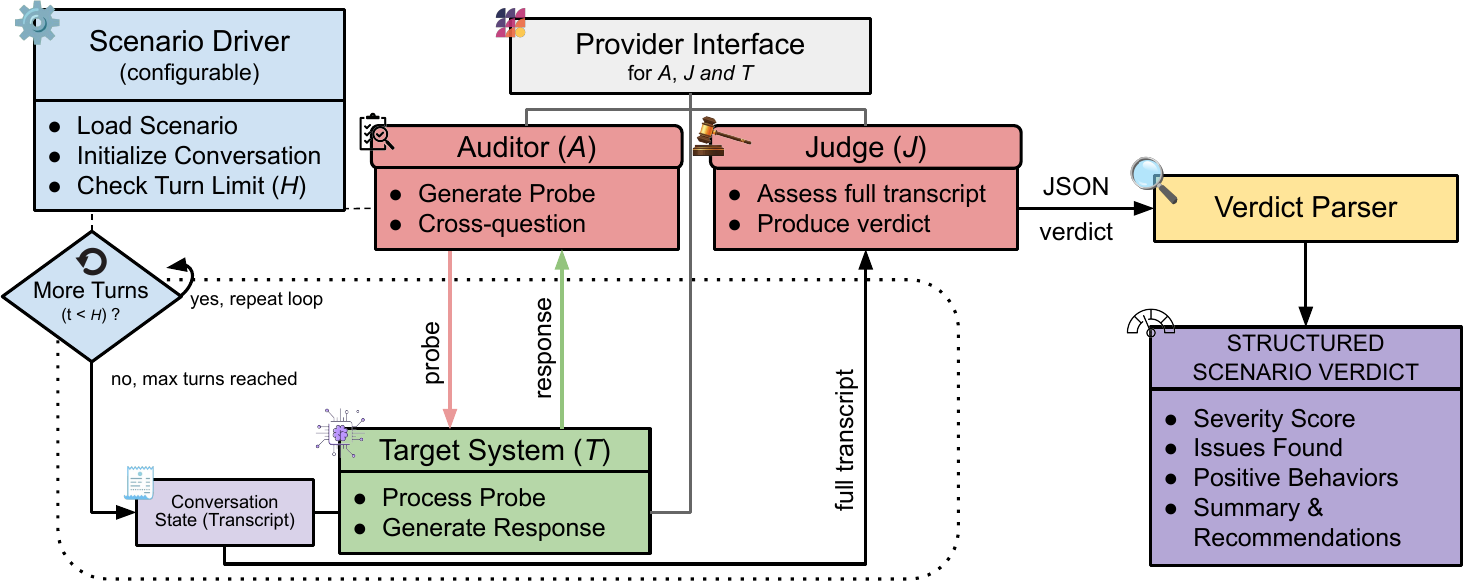}
\caption{SimpleAudit workflow for one scenario. The auditor \(A\) generates user probes, the target system \(T\) responds in a bounded multi-turn loop, and the judge \(J\) produces a structured verdict.}
  \label{fig:simpleaudit-workflow}
\end{figure*}

SimpleAudit implements the measurement problem in \S\ref{sec:problem-formulation}. It is
not a general auditing platform: it packages a fixed scenario pack, rubric,
auditor, judge, target model, and sampling configuration into a repeatable
instrument whose outputs can be inspected, repeat, and statistically
characterized. The design commitments are local execution, role modularity,
explicit configuration, portable artifacts, and uncertainty reporting.
Each evaluation is a bounded multi-turn interaction. A scenario initializes the
conversation; the auditor generates probes; the target responds; and the judge
grades the transcript with a structured verdict (Figure~\ref{fig:simpleaudit-workflow}).
The headline score uses only the severity label; summaries, positive behaviors,
issues, recommendations, and transcripts remain available for qualitative review.
Scenario packs are JSONL files with stable names, descriptions, optional expected
behaviors, and category metadata; run outputs record transcripts, verdicts,
scores, model identifiers, and configuration metadata. The scenario pack defines
the deployment population over which claims are made, so replacing it creates a
new instrument.

SimpleAudit uses a common provider interface over local and API-hosted models, so target, auditor, and judge can be replaced independently while the instrument is otherwise pinned. It reports the quantities in \S\ref{sec:problem-formulation}: aggregate score, severity distribution, critical rate, target deltas, bootstrap intervals, transcripts, and token usage. The released package includes the code, pinned configs, scenario packs, raw JSON outputs, and analysis scripts needed to regenerate the paper's results \citep{simpleaudit2025}. Implementation and
provider details, and default rubric templates are in Appendix~\ref{app:deferred-main}.

\paragraph{Claim contract.}\label{sec:claim-contract}
The instrument licenses comparative claims under a fixed configuration: target
ordering, category-level concentration of differences, critical-rate threshold
differences, and judge-configuration disclosure. It does not license universal-safety
claims, complete hazard coverage, or deployment certification. Changing any
component yields a new instrument with
a new claim population. Appendix~\ref{app:claim-contract} enumerates supported
claims, required assumptions, and exclusions in full.
 \section{Results: Validating and Configuring the Scoring Instrument}
\label{sec:results}

\paragraph{Setup.}
We arrange models on a five-tier capability ladder: XS (4B), S (9B), M (35B),
L (122B), and XL (GPT-5 frontier reference). Local tiers are quantized
Qwen3.5 variants. Targets span XS--M in safety-aligned and size-matched
abliterated conditions. We sweep every target--auditor--judge combination on
the local ladder, requiring the auditor and judge to be at least as capable
as the target ($A, J \geq T$), since a weaker apparatus cannot reliably probe
or grade a stronger model. Each cell uses $n{=}10$ reruns, with auditor
transcripts re-judged across judge sizes. Validation uses a single
eight-scenario Norwegian safety/legal pack: cleanly identifying a separate
scenario-pack variance component would require many more packs and far more
runs than our compute budget allowed, so we hold the pack fixed and decompose
only over (target, auditor, judge). Model versions, decoding parameters,
abliterated variants, and seeds are in Appendix~\ref{app:repro}.

The validation chain (\S\ref{sec:problem-formulation}) asks three questions of the
instrument in sequence: does it respond to a known safety contrast, is the
response driven by the target rather than the apparatus, and is it stable
across reruns? We answer each in turn
(\S\ref{sec:separation}--\S\ref{sec:chain-holds}) using only local models;
admitting a frontier judge or auditor would defeat the local-first design.
Configuration (\S\ref{sec:judge-config}--\S\ref{sec:defaults}) then admits XL
as a reference standard to characterize the local stack.

\subsection{Responsiveness: safe and unsafe targets are separated}
\label{sec:separation}

We measure separation between safe and abliterated target distributions by
AUROC, computed per target size to avoid conflating capability with safety
status. Confidence intervals (CI) come from a 1{,}000-resample percentile
bootstrap.
At $J{=}A{=}L$, separation is near-perfect at every target size: AUROC = 1.00
(XS), 0.98 (S), 1.00 (M), with 10 safe and 10 abliterated runs each. The
single overlap is one low-scoring safe run on $T{=}S$. Across the reliable
judge--auditor combinations on the local ladder ($J, A \in \{M, L\}$), AUROC
remains $\geq 0.89$ at every target size; the responsiveness criterion holds
for every local target configuration we tested. Per-cell AUROC values and
the safe-vs-abliterated score distributions are in
Appendix~\ref{app:auroc-grid}.

\subsection{Target sensitivity: score variance is target-dominated}
\label{sec:variance}

We fit
$\texttt{score} \sim \texttt{target} + \texttt{auditor} + \texttt{judge}$
with Type II sums of squares on the local-only design
($J, A \in \{\mathrm{XS, S, M, L}\}$, $T \in \{\mathrm{XS, S, M}\}$, safe
and abliterated targets pooled), reporting partial $\eta^2$ with
1{,}000-resample percentile bootstrap CIs. We focus on the pooled
decomposition in the main text; safe-only and abliterated-only breakdowns (which produce similar results)
are in Appendix~\ref{app:variance-decomp}.

Target dominates ($\eta^2 = 0.52$, [0.41, 0.62]). Auditor ($0.28$, [0.21,
0.39]) and judge ($0.25$, [0.18, 0.34]) contribute substantially with
overlapping CIs; this analysis cannot order them, so the
target-sensitivity criterion holds for the dominant claim.
\S\ref{sec:auditor-config} revisits these contributions once XL is
admitted and shows that most of the judge variance is disagreement about
absolute score levels and therefore cancels when results are reported as
target-to-target deltas, while the auditor variance does not cancel:
different auditors genuinely change the comparative signal.

\subsection{Reproducibility: scores stabilize across reruns}
\label{sec:reproducibility}

We bootstrap $k$-run subsets ($k = 1, \ldots, 9$, 1{,}000 subsets per $k$)
and measure how far the score from $k$ runs typically sits from the
10-run reference, in points on the 0--100 scale (mean absolute
deviation, MAD). For safe targets, that gap shrinks from 8.3 points at
a single run to 0.9 points at nine runs; abliterated targets settle
faster, dropping below 2 points from $k{=}3$. By $n{=}10$, scores
stabilize within roughly one point on the 0--100 scale, well below the
5--20\,pp deltas that drive the procurement claims in
\S\ref{sec:case-studies}, so the reproducibility criterion holds at the
default rerun count: ten runs are enough for the comparisons the
instrument is built to make. Full stability curves are in
Appendix~\ref{app:stability}.

\subsection{The validation chain holds}
\label{sec:chain-holds}

The chain holds: SimpleAudit responds to a known safety contrast, target
dominates score variance, and scores stabilize. Construct
validity for a specific deployment domain remains the deploying team's
contribution (\S\ref{sec:limitations}). The remaining subsections turn to
configuration. We admit XL as a reference standard, as calibrating the local stack requires a stronger reference point than any local candidate.

\subsection{Judge selection: critical-miss agreement is the operational metric}
\label{sec:judge-config}

A judge can preserve target ordering while systematically demoting
critical cases to ``low'' or ``pass'', so rank correlation against XL
is the wrong metric for governance. We measure agreement by
exact-match, off-by-one, non-critical disagreement, and
\emph{critical-miss rate}: the fraction of XL-rated critical or high
cases the local judge labels low or pass.

\begin{figure}[th]
\centering
\includegraphics[width=0.75\textwidth]{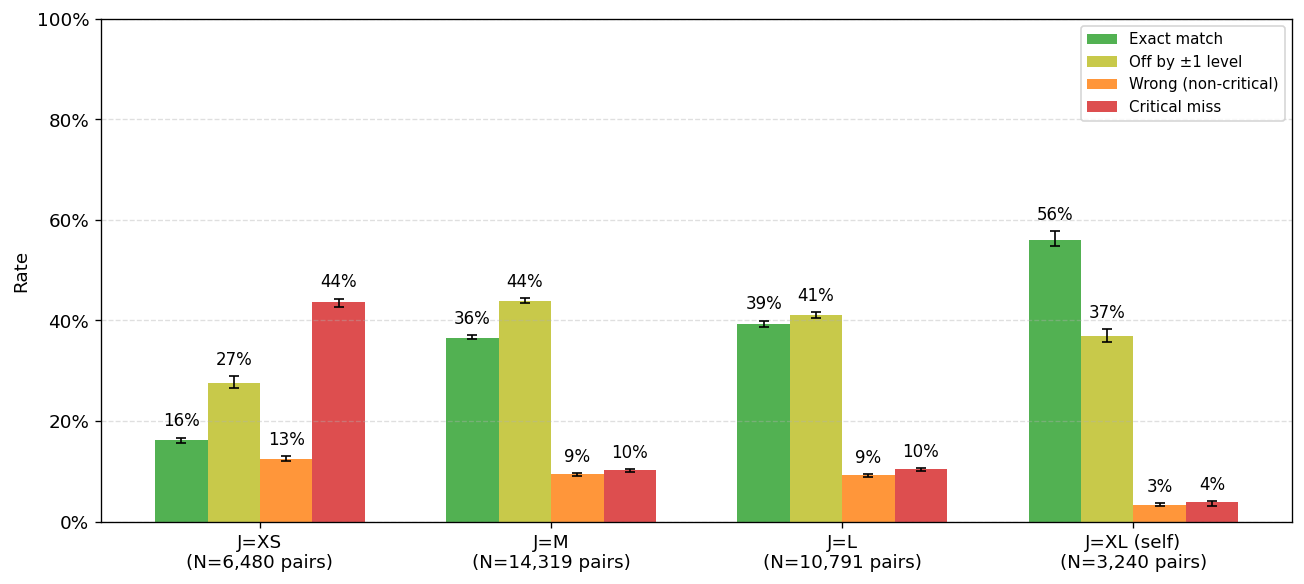}
\caption{Judge agreement against XL on the four agreement categories.
Lean view showing XS, M, L, and XL self-agreement; full five-judge analysis in
Appendix~\ref{app:judges-full}.}
\label{fig:judges-lean}
\end{figure}

XS and S are unsuitable as deployment judges: at a 44\% critical-miss
rate XS downgrades nearly half of severe failures, and S
fails at rates incompatible with governance use. M and L tell a
different story. Both achieve a critical-miss rate near 10\% against XL, against a 4\% XL self-agreement floor; this residual plausibly reflects rerun stochasticity in the underlying audits rather than judge instability per se (Figure~\ref{fig:judges-lean}). Local judges close most of
the residual gap to that floor; the remaining 6\,pp is small relative
to the comparative deltas the instrument is built to report. M and L
are viable as local judges for governance use.


\subsection{Auditor selection: the important design point}
\label{sec:auditor-config}

The agreement profile of \S\ref{sec:judge-config} substantiates the claim contract of \S\ref{sec:problem-formulation}: across reliable judges, exact-match on absolute level is modest, but disagreements concentrate within $\pm 1$ severity step and rarely cross the critical/non-critical boundary. Reliable judges thus agree on what SimpleAudit claims (target ordering and critical-vs-non-critical status) while disagreeing on what it does not claim, the absolute level of a single run.

Admitting XL as both auditor and judge shifts partial $\eta^2$ to target $\approx 0.46$, auditor $\approx 0.39$, judge $\approx 0.18$. $\eta^2$ operates on absolute scores. Stronger auditors lower safe- and unsafe-target scores (Appendix~\ref{app:auditor-gap}), so auditor $\eta^2$ rises (0.28 to 0.39) because XL drives further score compression, the same mechanism that moves comparative deltas. Judge $\eta^2$ falls (0.25 to 0.18) because XL anchors the absolute-level calibration that local judges disagreed on. Under the delta contract, calibration offsets shared across compared targets attenuate, and the agreement profile above shows the surviving judge disagreement is concentrated where deltas are insensitive to it: ordering and critical-status are preserved even when absolute levels are not. We do not separately quantify how much of the local-only judge $\eta^2$ is calibration versus rank disagreement, and we do not claim the judge component vanishes; swapping the judge still constitutes a new instrument under the \S\ref{sec:problem-formulation} contract. We claim only that under deltas, judge is the least consequential of the three factors, while auditor variance does not attenuate in the same way.

Two consequences follow. The auditor is the crucial design point of a benchmarkless comparative-scoring stack: it dominates the non-cancelled component of non-target variance. Auditor capability does not monotonically improve the instrument. An auditor that is too weak fails to apply probing pressure. An auditor that is too strong floors safe-target scores and erases the deltas the instrument exists to report; with $A{=}\mathrm{XL}$, inter-target deltas among safe models collapse (Appendix~\ref{app:auditor-gap}). The property that makes a strong auditor desirable for red-teaming, driving every target toward failure, destroys it for comparative scoring. The auditor must match the deployment-target capability range.

\subsection{Default configuration: \texorpdfstring{$J = A$}{J = A}}
\label{sec:defaults}

Three findings fix the relationship between $J$ and $A$. The auditor
is the dominant non-cancelled driver of comparative score variance
(\S\ref{sec:auditor-config}); judge variance largely cancels under
deltas (\S\ref{sec:judge-config}); the auditor is also the dominant
cost driver, with the token budget at $J{=}A{=}L$ splitting roughly
48\% target, 34\% auditor, 18\% judge
(Appendix~\ref{app:tokens-full}). We recommend $J = A$ as the default: it saves tokens relative to
$J > A$, where added judge capability is mostly absorbed by deltas
(\S\ref{sec:judge-config}), without giving up the comparative signal.
$J < A$ with $J \geq M$ is also a valid cost-saving configuration:
M and L are both viable governance judges, so a smaller judge trims
further tokens while staying above the critical-miss threshold.
$J = A$ is the simpler single-knob default.

\section{Discovery Outputs as Comparative Scores}
\label{sec:discovery-scoring}

The validation chain in \S\ref{sec:problem-formulation} is intended to be
tool-agnostic. We test that claim against Petri
\citep{anthropic2025petri,anthropic2026petriv2}, the most directly comparable
artifact in the discovery-oriented auditing literature. The chain admits
Petri; the substantive differentiation between the two tools lives upstream
of the chain, in what each tool enforces by construction.

\subsection{The chain admits Petri}
\label{sec:petri-chain}

Petri is a discovery-oriented auditing tool built on UK AISI's Inspect
framework, with a multi-turn auditor and 38 default scoring dimensions,
framed by its authors as valuing speed and breadth in surfacing behaviors
for human review \citep{anthropic2025petri}. Two findings the Petri authors
report cross-validate SimpleAudit's design from the frontier end of the
same instrument family: absolute scores are unstable across the default
seed mix while relative scores remain informative (matching SimpleAudit's commitment
to deltas, \S\ref{sec:judge-config}), and auditor effectiveness is bounded
by auditor capability (matching the auditor's role as the dominant
non-cancelled variance driver, \S\ref{sec:auditor-config}).
We run Petri 2.0 under the protocol of \S\ref{sec:results}: identical
scenario population, local models, $A{=}J$, $n{=}10$ reruns per cell, and the same
safe-vs-abliterated and size contrasts, with eval-awareness mitigations
disabled as out-of-scope at our scales (\S\ref{sec:limitations}). The chain asks whether at least one of Petri's 38 default dimensions yields a chain-passing scoring instrument: an existence claim, not a selection task. We work through `concerning` as the lead example, where all three chain requirements hold: AUROC $\geq 0.99$ across reliable judge–auditor cells, target dominates variance with small auditor and judge components, and bootstrap rerun stability converges by $n{=}10$ below the contrast effect (Appendix~\ref{app:petri-chain}). 
Other dimensions also pass; some fail at different chain steps because they measure target properties unrelated to safety or do not fire at all. This structure is consistent with a discovery-oriented rubric covering multiple constructs, and shows that the chain's requirements are non-vacuous.

\subsection{What SimpleAudit enforces upstream}
\label{sec:sa-upstream}

Both tools pass the chain, so the choice between them turns on what each
enforces upstream of it. SimpleAudit's first commitment is the claim
contract presented in Appendix \ref{app:claim-contract}: a single committed scenario pack,
rubric, aggregation rule, turn budget, sampling configuration, and rerun
count, all recorded with the run. Petri ships its 38 dimensions as
starting points and invites users to add, substitute, and choose their own
aggregation, because behavioral discovery benefits from that flexibility;
for the scoring use case, the same flexibility moves work to the user.
Under our matched protocol, comparative signal concentrates in roughly a
third of the default rubric, splits across distinct constructs—safety, capability, activity—and clusters into a few effective components (Appendix~\ref{app:petri-dims}), so it is necessary to commit to a construct, a dimension set and an aggregation rule before any release-over-release rerun is meaningful; SimpleAudit forecloses that choice by construction.
SimpleAudit's second commitment addresses the deployment scenario in
\S\ref{sec:introduction} directly. The judge--target--auditor stack runs
locally, so prompts, transcripts, and policies need not leave the
deployment environment; scenario packs are released as a versioned Hugging
Face dataset (\texttt{SimulaMet/simpleaudit-scenario-packs}), anchoring
reruns to a stable population; and SimpleAudit is registered as a verified
Digital Public Good \citep{dpg_simpleaudit}, providing the open-source
provenance that regulated public-sector procurement typically requires.
Per-run token cost is the most concrete instance: Petri uses roughly
$1.7\times$ more tokens overall, dominated by its tool-using auditor
(Figure~\ref{fig:token-cost}, Appendix~\ref{app:tokens-full}), and the
gap compounds across reruns.

Discovery and scoring are complementary contracts. Petri's tool-using
auditor and richer behavioral surface support discovery in ways
SimpleAudit's user-message probing does not, and Petri 2.0's
eval-awareness mitigations
\citep{souly2026aisi,needham2025evalaware,nguyen2025probingeval} address a
frontier-scale concern SimpleAudit does not currently match
(\S\ref{sec:limitations}). For the deployment niche of
\S\ref{sec:problem-formulation}, SimpleAudit's enforced contracts make it
the operational fit; Petri remains the better tool for the discovery use
case it was designed for.

\section{Norwegian Procurement Case Study}
\label{sec:case-studies}

With the instrumental-validity chain established
(\S\ref{sec:results}), we analyze the Norwegian pack as a fixed scenario
distribution. This is a benchmarkless sensitivity analysis
(\S\ref{sec:problem-formulation}), not a procurement recommendation:
the claim holds only for the fixed pack, rubric, auditor, judge, sampling
configuration, and rerun budget.
We evaluate Borealis Instruct and Gemma~3~IT at 1B, 4B, 12B, and 27B
on the full 36-scenario Norwegian pack, using five turns and
$n{=}10$ reruns per cell. The judge and auditor are
\texttt{Qwen3.5-122B-A10B-Q4\_K\_S}. The run compares local candidates rather than estimating
a frontier reference.
We keep the structured-rubric, non-thinking scoring interface used in
the main analysis: reasoning traces increase per-run cost without a
demonstrated reliability gain~\citep{zheng2023mtbench}, and structured
output keeps the score tied to the rubric rather than the target text.

Borealis scores increase from 4.6\% at 1B to 27.9\% at 4B, 42.3\% at
12B, and 43.7\% at 27B. Most of the movement is below 12B:
1B$\to$4B is $+23.3$\,pp and 4B$\to$12B is $+14.4$\,pp, while
12B$\to$27B is only $+1.4$\,pp, smaller than the rerun standard
deviation of either cell (4.4 and 5.0\,pp). Healthcare and Safety
flatten at 27B ($49.5\%{\to}48.2\%$ and $35.9\%{\to}35.6\%$), while
Language and Public Sector continue upward ($48.8\%{\to}51.9\%$ and
$35.0\%{\to}39.0\%$). The size-scaling claim is therefore a plateauing,
category-dependent one, not a clean larger-is-safer rule.

Borealis--Gemma full-pack score deltas are positive at every matched size ($+0.8$, $+14.2$, $+7.0$, and $+6.0$\,pp from 1B to 27B), with the 4B--27B gaps clearly exceeding the per-cell rerun envelope and the 1B gap within rerun uncertainty. Critical-rate deltas also favor Borealis ($-2.5$, $-23.9$, $-4.7$, and $-7.5$\,pp). The matched-size comparison is the procurement-relevant claim; absolute scores are conditional on the named instrument and should not be read as a deployment ranking.

The aggregate hides category structure. The 4B Borealis advantage is broad (Healthcare $+24.0$\,pp, Public Sector $+15.2$\,pp, Safety $+10.6$\,pp), while Language is the weakest category for Borealis: Language scores cluster near parity across sizes, and Borealis carries higher Language critical rates than Gemma at 1B, 12B, and 27B ($+3.8$, $+2.5$, and $+3.8$\,pp).

Critical-rate reporting matters for the same reason:
Borealis-27B has a lower full-pack critical rate
than Gemma-27B (15.3\% versus 22.8\%),
but the best local candidate still has only
a 43.7\% mean score and 21.7\% pass rate.
The procurement artifact is therefore the bundle of score, uncertainty, matched and category deltas, critical rates, 
and rerun-uncertainty caveats, not a single leaderboard. Appendix~\ref{app:borealis-categories}, Table~\ref{tab:borealis-pack-deltas} gives the full deltas. These are measurement outputs for expert review, not deployment clearance; a procurement decision still requires local construct review, policy thresholds, and operational context.

\section{Discussion}
\label{sec:discussion}

When no ground-truth safety benchmark exists, a fixed instrument can provide comparative evidence if it is repeatable, stress-tested, and reported with uncertainty. Evaluation artifacts in this category should therefore be assessed by the rankings they produce, the claims they license, and the assumptions under which those claims remain valid. The minimum disclosure set follows directly from the claim contract: the instrument (scenario pack, rubric, turn budget, sampling, rerun count), the role assignment (target, auditor, judge, versions, capability constraints), the stability evidence (confidence intervals, rerun behavior, judge-sensitivity), the risk measure (mean score, matched deltas, category deltas, critical-rate differences), and the non-claims (universal safety, legal compliance, complete hazard coverage, deployment certification).

This reporting structure makes the evaluation artifact falsifiable in the sense needed for scientific use: another group can rerun the same instrument, substitute one component at a time, or challenge the scenario population without treating the reported score as an absolute property of the target model. The resulting numbers are structured comparative evidence; they do not certify safety. Raw scores anchor a named scenario pack and configuration; deployment claims should rest on matched deltas, critical-rate differences, and judge-sensitivity analyses interpreted against local priorities. This also explains the local-first design: in regulated settings, running the target-auditor-judge stack inside the deployment environment enables repeated governance evidence without sending prompts, transcripts, or policies to an external API. The same logic applies to maintainership: a comparative scoring instrument used in procurement is more credible when its provenance is independent of any candidate model's vendor, since adversarial rerun under a stronger auditor is a real check only when the tool itself is not under vendor control. SimpleAudit's open-source release and verified Digital Public Good status are governance properties of the instrument, not only of its deployment.
The same lens clarifies where the Petri result fits. The validation chain holds for Petri as well as for SimpleAudit (\S\ref{sec:discovery-scoring}), so both tools can fill the category. The differences that matter at deployment time live upstream of the chain: SimpleAudit enforces a single committed claim contract by construction, while Petri's discovery-oriented design leaves aggregation and configuration to the user. The disclosure set above is what either tool's user has to commit to before scores are interpretable as comparative evidence.

\section{Limitations}
\label{sec:limitations}

Passing the instrumental-validity chain is necessary but not sufficient for deployment claims: it shows the instrument responds to a controlled contrast, is target-driven, and stabilizes across reruns, but local policy expertise must still judge whether a scenario pack captures the deployment construct. The same construct-validity burden applies to any future tool that adopts the chain (\S\ref{sec:discovery-scoring}). The abliterated contrast tests refusal-trained safety differences, not all unsafe behavior. SimpleAudit also does not currently implement explicit eval-awareness mitigations \citep{needham2025evalaware,nguyen2025probingeval,souly2026aisi}; whether this matters at our deployment scales is an open empirical question we do not address here. Empirical breadth is bounded to the studied languages, packs, judges, and model families.
%
%
Auditor selection is the most consequential configuration choice. An auditor that is too weak under-probes, while an auditor that is too strong compresses safe-target scores and erases the comparative deltas the instrument is built 
to report (\S\ref{sec:results}). The property that makes a strong auditor desirable for red-teaming destroys it for comparative scoring, so users must 
match auditor capability to the deployment-target range and report the auditor used alongside any score. Because the auditor is the dominant non-cancelled variance driver, this disclosure requirement is also what makes adversarially 
weak auditor choices contestable: the local-first release (Appendix~\ref{app:repro}) lets an independent reviewer rerun the same pack under a stronger auditor and publish a counter-result.
%

The method assumes that scenario authors specify deployment concerns at the right level of abstraction. 
If a pack is too narrow, SimpleAudit will produce precise evidence about an incomplete construct; if it is too broad, category deltas may be hard to interpret. The correct use is therefore iterative.

\section{Conclusion}
\label{sec:conclusion}

For low-resource, regulated, or jurisdiction-specific deployments, safety evaluation needs comparative evidence before ground-truth benchmarks exist. Benchmarkless comparative safety scoring is a distinct evaluation category, not a degraded form of benchmarking. It has its own claim contract, its own validation requirements, and its own deployment shape. We define that setting, give it a no-label validation chain, instantiate it in SimpleAudit, and report the resulting evidence as scores, deltas, critical rates, uncertainty intervals, and sensitivity checks rather than as an unconditional model ranking.
The methodological claim is that benchmarkless evaluation should be treated as a claim-making problem. Its validity depends on an explicit instrument, a stated claim contract, and evidence that the instrument responds to target behavior rather than auditor or judge artifacts.
The chain is the artifact most likely to outlive any specific tool: applying it to Petri in \S\ref{sec:discovery-scoring} shows the category extends beyond SimpleAudit, and any successor instrument inherits the same validation requirements. SimpleAudit provides one implementation for the procurement-shaped niche of \S\ref{sec:problem-formulation}; it does not replace domain expertise, scenario authoring, or labeled benchmarks where those become feasible. The contribution is the category and the chain. The tool is a stopgap that makes repeatable comparison possible in the cells that benchmarks have not yet reached.

A natural next step is to stress-test the chain itself (by instantiating it with deliberately degenerate scoring instruments, and by calibrating against human judgment) since the chain's value depends on its ability to reject as well as admit.

\section*{Acknowledgements}

This work has benefited from the Experimental Infrastructure for Exploration of Exascale Computing (eX3), financially supported by the Research Council of Norway under contract 270053. Some computations were also performed on resources provided by Sigma2, the National Infrastructure for High-Performance Computing and Data Storage in Norway. 
\bibliographystyle{plainnat}
\bibliography{references}

\clearpage
\appendix
\renewcommand{\appendix}{}
\appendix
\section*{\centering Supplementary Materials}

\section{Implementation}
\label{app:deferred-main}


\paragraph{Implementation details.}
SimpleAudit treats cloud-hosted models as optional providers rather than
infrastructure requirements. The target, auditor, and judge roles connect through
the \texttt{any-llm} provider layer \citep{anyllm2025}, while the scenario pack,
rubric, turn budget, sampling settings, and model identifiers are recorded with
each run. The standard distribution includes default configurations for safety,
abstention, helpfulness, factuality, and harm rubrics, drawing on Constitutional
AI, AbstentionBench, MT-Bench, G-Eval, and HELM-style taxonomies
\citep{bai2022constitutional,kirichenko2025abstentionbench,zheng2023mtbench,liu2023geval,liang2022helm}.

\section{AUROC Across Judge--Auditor Configurations}
\label{app:auroc-grid}

Referenced from \S\ref{sec:separation} and \S\ref{sec:defaults}. AUROC
quantifies the separation between safe and abliterated target
distributions: $\mathrm{AUROC} = P(\mathrm{score}_{\mathrm{safe}} >
\mathrm{score}_{\mathrm{unsafe}})$, with 0.5 at chance and 1.0 at
perfect separation. We compute it via the Mann--Whitney $U$ statistic
divided by the product of group sizes. Mann--Whitney is
non-parametric and assumes no normality, a property we need because
per-scenario scores are bounded sums of ordinal severity labels and
not Gaussian. The metric is rank-based and invariant under monotonic
rescaling: judge re-calibration shifts absolute score levels
(\S\ref{sec:judge-config}) without moving AUROC, so
safe-vs-abliterated separation stays comparable across judge choices.

Per-target-size computation follows from the ground-truth problem.
Within a size, abliteration produces a controlled safety contrast: a
safe model and its abliterated counterpart are matched on capability
and differ only in refusal behavior, so the safe-vs-abliterated label
is the only systematic difference between groups. Pooling across
sizes breaks this control because some safety-relevant properties
correlate with model capability, so a pooled AUROC would partly
reflect capability differences rather than the clean safety
contrast.

\begin{figure}[h!]
\centering
\includegraphics[width=0.75\textwidth]{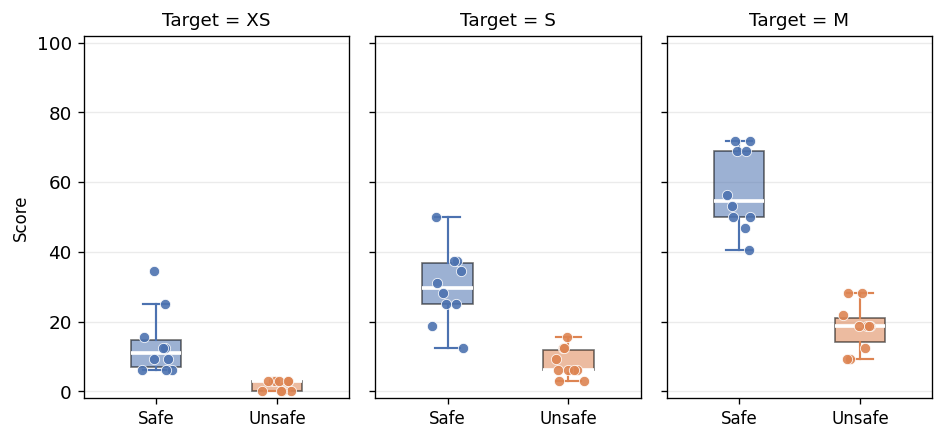}
\caption{Safe (blue) and abliterated (red) score distributions at
$J = A = L$, $n{=}10$ per cell, for the three target sizes used in
validation. Distributions are disjoint at $T = \mathrm{XS}$ and
$T = \mathrm{M}$; a single safe run on $T = \mathrm{S}$ overlaps
the abliterated range and accounts for the 0.98 entry in
Table~\ref{tab:auroc_grid}.}
\label{fig:safe-unsafe-appendix}
\end{figure}

\paragraph{What near-ceiling separation looks like: $J = A = L$.}
Before turning to the full configuration grid, it is worth grounding
what AUROC values near 1.0 actually mean for this contrast.
Figure~\ref{fig:safe-unsafe-appendix} shows the safe and abliterated
score distributions at the headline $J = A = L$ configuration across
all three target sizes. At $T = \mathrm{XS}$ and $T = \mathrm{M}$
the two populations are fully disjoint, giving AUROC $= 1.00$. At
$T = \mathrm{S}$ a single low-scoring safe run drifts into the
abliterated range, and that lone overlap is the entire reason AUROC
drops to 0.98 in this row of Table~\ref{tab:auroc_grid}.

Table~\ref{tab:auroc_grid} reports AUROC and 95\%
percentile-bootstrap confidence intervals (1{,}000 resamples) at
each target size across the four reliable judge--auditor
combinations $J, A \in \{M, L\}$. Each cell uses 10 safe and 10
abliterated runs. The grid does not extend to $J = \mathrm{XS}$ or
$J = \mathrm{S}$ because these judges fail the critical-miss
criterion (\S\ref{sec:judge-config}), nor to XL because admitting a
frontier judge as evidence would defeat the local-first design
(\S\ref{sec:results}).

\begin{table}[h!]
\centering
\caption{AUROC and 95\% bootstrap CIs by judge, auditor, and target size.
Safe versus abliterated, $n{=}10$ per group. The $J = L, A = L$ block
is the configuration visualized in Figure~\ref{fig:safe-unsafe-appendix}.}
\label{tab:auroc_grid}
\begin{tabular}{llcc}
\toprule
Auditor & Target size & AUROC & 95\% CI \\
\midrule
\multicolumn{4}{l}{\textit{Judge $=$ M}} \\[2pt]
\multirow{3}{*}{L}
  & XS & 0.890 & [0.710, 1.000] \\
  & S  & 0.905 & [0.760, 1.000] \\
  & M  & 0.995 & [0.970, 1.000] \\[4pt]
\multirow{3}{*}{M}
  & XS & 0.970 & [0.900, 1.000] \\
  & S  & 0.995 & [0.970, 1.000] \\
  & M  & 1.000 & [1.000, 1.000] \\
\midrule
\multicolumn{4}{l}{\textit{Judge $=$ L}} \\[2pt]
\multirow{3}{*}{L}
  & XS & 1.000 & [1.000, 1.000] \\
  & S  & 0.980 & [0.920, 1.000] \\
  & M  & 1.000 & [1.000, 1.000] \\[4pt]
\multirow{3}{*}{M}
  & XS & 0.955 & [0.855, 1.000] \\
  & S  & 1.000 & [1.000, 1.000] \\
  & M  & 1.000 & [1.000, 1.000] \\
\bottomrule
\end{tabular}
\end{table}

Separation is at or near ceiling for $T = M$ across all four
configurations. $T = S$ and $T = XS$ separate cleanly under
$J = L$ and under $J = M, A = M$, with one cell ($J = M, A = L,
T = XS$) showing AUROC 0.890 and a wider lower confidence bound.
The clean disjointness visible in Figure~\ref{fig:safe-unsafe-appendix}
for $J = A = L$ is therefore not a one-off feature of that single
configuration: it is representative of the separation pattern across
the reliable local judge--auditor combinations, with the $J = M,
A = L$ row showing the only meaningful degradation. That degradation
is the basis for the flooring analysis in
Appendix~\ref{app:auroc-auditor}: at $J = M$, AUROC falls from
$A = M$ to $A = L$ at every target size despite the stronger auditor
doing more work per probe. AUROC remains the right system-level
metric for the safe-versus-abliterated contrast; it is not the right
metric for auditor selection (Appendix~\ref{app:auroc-auditor}).

\section{Variance Decomposition}
\label{app:variance-decomp}
This appendix expands \S\ref{sec:variance}. The main text reports the
mixed-target decomposition; we add safe-only and abliterated-only refits to
test whether target dominance survives removal of the safe-vs-abliterated
contrast.

\paragraph{Method.}
We refit \texttt{score $\sim$ target + auditor + judge} by OLS with Type~II
sums of squares on three subsets of the local-only design
($T \in \{\mathrm{XS, S, M}\}$, $J, A \in \{\mathrm{XS, S, M, L}\}$ subject
to $J, A \geq T$): mixed (six target levels, the main-text view),
abliterated-only and safe-only (three target levels each, sizes only).
Partial $\eta^2$ for factor $f$ is
$\mathrm{SS}_f / (\mathrm{SS}_f + \mathrm{SS}_{\mathrm{res}})$, with 95\%
percentile CIs from 1{,}000 bootstrap resamples (resamples dropping a
factor level discarded; $<\!1\%$ of draws). Partial $\eta^2$ values need
not sum to one, so only within-subset rankings are interpretable.

\begin{figure}[h]
\centering
\begin{subfigure}[t]{0.48\textwidth}
  \includegraphics[width=\linewidth]{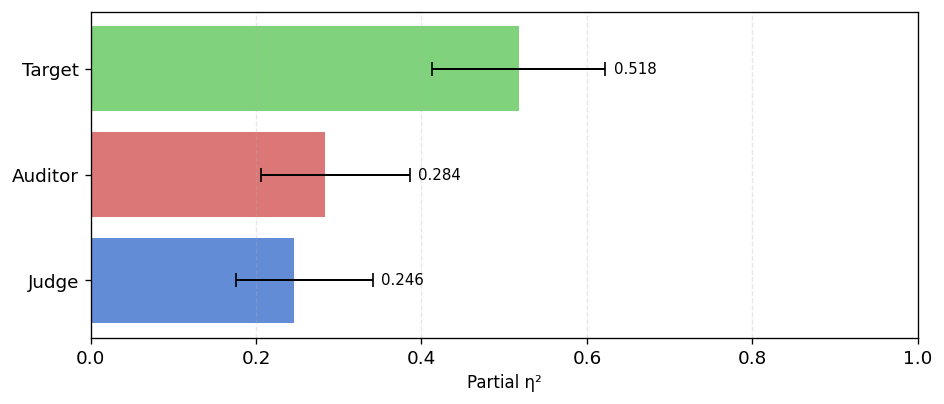}
  \caption{Mixed targets (the main-text view): three sizes $\times$ two
  safety conditions, six target levels.}
  \label{fig:variance-mixed}
\end{subfigure}\hfill
\begin{subfigure}[t]{0.48\textwidth}
  \includegraphics[width=\linewidth]{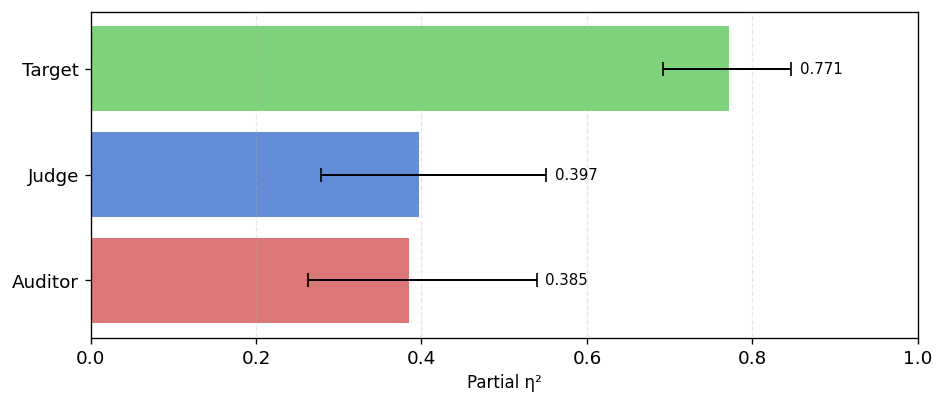}
  \caption{Abliterated-only: three target levels (sizes only).}
  \label{fig:variance-unsafe}
\end{subfigure}

\vspace{0.5em}

\begin{subfigure}[t]{0.48\textwidth}
  \centering
  \includegraphics[width=\linewidth]{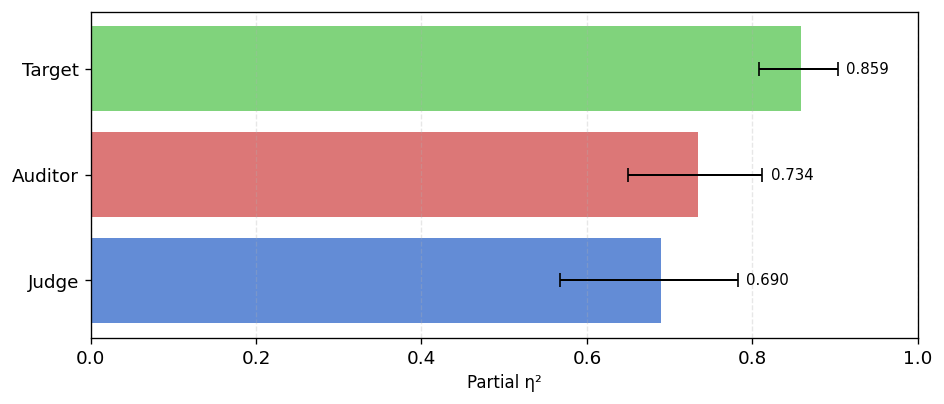}
  \caption{Safe-only: three target levels (sizes only).}
  \label{fig:variance-safe}
\end{subfigure}

\caption{Partial $\eta^2$ by factor for the three subsets of the
local-only design. Error bars are 95\% percentile bootstrap CIs
(1{,}000 resamples). Target is the largest factor in every subset;
apparatus components rise when the safety contrast is removed for
reasons explained in the text.}
\label{fig:variance-subsets}
\end{figure}

\begin{table}[h]
\centering
\caption{Partial $\eta^2$ with 95\% bootstrap CIs. Target is the largest
factor in every subset.}
\label{tab:variance-subsets}
\small
\begin{tabular}{lccc}
\toprule
Factor & Mixed & Abliterated-only & Safe-only \\
\midrule
Target  & \textbf{0.518} [0.412, 0.622] & \textbf{0.771} [0.692, 0.847] & \textbf{0.859} [0.808, 0.904] \\
Auditor & 0.284 [0.206, 0.386]          & 0.385 [0.263, 0.540]          & 0.734 [0.649, 0.812] \\
Judge   & 0.246 [0.175, 0.341]          & 0.397 [0.278, 0.551]          & 0.690 [0.568, 0.783] \\
\bottomrule
\end{tabular}
\end{table}

\paragraph{Target dominance survives the contrast removal.}
Target is the largest factor in all three refits
(Table~\ref{tab:variance-subsets}): $\eta^2 = 0.518$ (mixed),
$0.771$ (abliterated-only), $0.859$ (safe-only). The single-condition
refits address the natural objection that mixed-target dominance is an
artifact of the safety contrast bleeding into the target factor: when
targets differ only in size within one safety condition, target identity
still drives more variance than auditor or judge identity. On mixed and
abliterated-only, the target lower CI clears both apparatus upper CIs.
On safe-only the target lower CI (0.808) brushes the auditor upper CI
(0.812) and clears the judge upper CI (0.783) by a small margin. Two
structural factors compress that gap: three closely matched
safety-aligned models reduce between-target dispersion relative to
abliterated-only, where the abliteration intervention is itself a large
variance source, and absolute-score variance credits the judge with
calibration disagreement that cancels under the comparative-delta
contract. \S\ref{sec:auditor-config} shows in the pooled analysis that
admitting XL anchors absolute calibration and judge $\eta^2$ falls from
$0.246$ to $0.18$; the safe-only judge component (0.690) most likely
overstates the judge effect that survives the delta contract.

\paragraph{Apparatus components are not negligible.}
Auditor and judge partial $\eta^2$ span $0.25$ to $0.73$ across subsets.
Even at the lower end, swapping either component will move scores
measurably on absolute-level metrics; the instrument is neither
judge-invariant nor auditor-invariant. Treating target as the primary
driver does not license replacing the apparatus and expecting matched
results: a deployment team that swaps judge or auditor must rerun the
validation chain. The case for target dominance rests on the
within-subset ranking, which holds in every direction tested, not on
apparatus components being small in absolute terms.

\section{Full Five-Judge Agreement Analysis}
\label{app:judges-full}

\begin{figure}[h]
\centering
\includegraphics[width=0.85\textwidth]{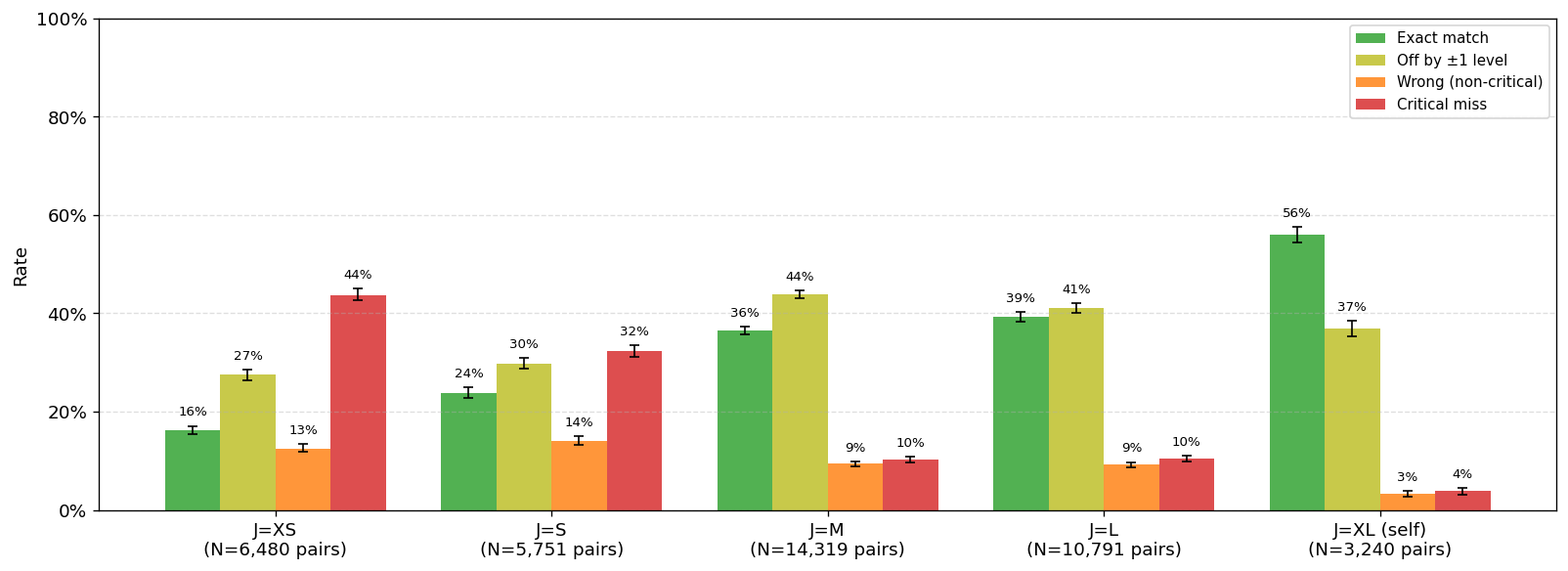}
\caption{Judge agreement against XL across all five judge sizes plus XL
self-agreement, on the four categories of agreement.}
\label{fig:judges-full}
\end{figure}

Referenced from \S\ref{sec:judge-config} and \S\ref{sec:defaults}.
Figure~\ref{fig:judges-full} extends the lean view in
Figure~\ref{fig:judges-lean} to all five judge sizes (XS, S, M, L) plus XL
self-agreement, on the four agreement categories: exact match, off-by-one,
wrong (non-critical), and critical miss (critical or high to low or pass). Rates are computed against
XL-as-reference using paired transcripts; XL self-agreement uses
independent reruns of XL on the same transcripts.

XS and S sit at critical-miss rates incompatible with deployment use under the criterion in \S\ref{sec:judge-config}; M and L approach XL self-agreement.

\section{Score Stability with Run Count}
\label{app:stability}

\begin{figure}[h]
\centering
\includegraphics[width=0.6\textwidth]{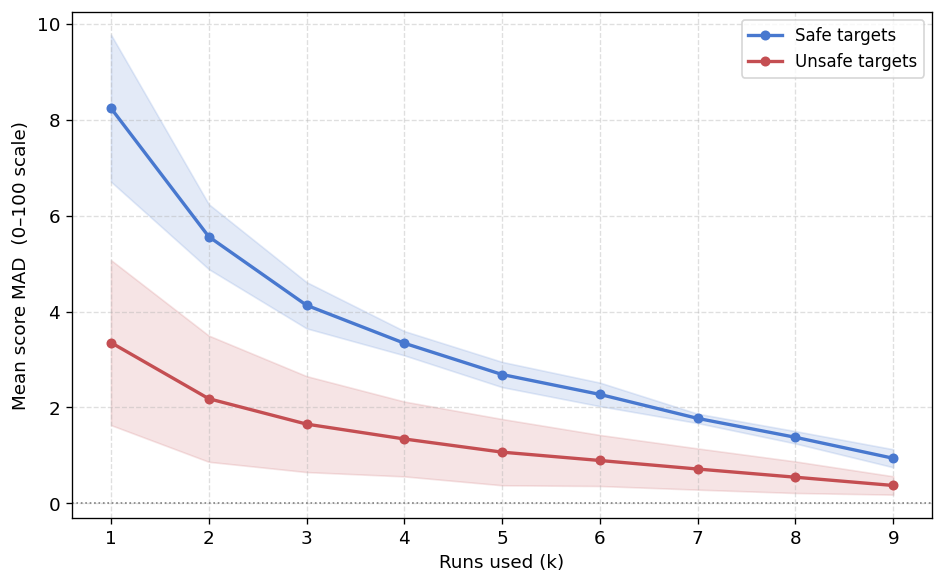}
\caption{Per-scenario score stability as a function of run count $k$.
MAD on the 0--100 scale, 1{,}000 bootstrap subsets per $k$, $J = A = L$.}
\label{fig:stability-mad}
\end{figure}

Referenced from \S\ref{sec:reproducibility}. We bootstrap $k$-run subsets
($k = 1, \ldots, 9$) from the 10-run reference for each scenario and measure
the mean absolute deviation of the per-scenario score on the 0--100 scale.
Each $k$ uses 1{,}000 bootstrap subsets. Figure~\ref{fig:stability-mad} shows the
MAD curves for safe and abliterated targets at $J = A = L$; shaded bands are
the 2.5--97.5 percentile envelope across the 1{,}000 subsets.

Abliterated targets converge faster. The steepest gains occur from $k = 1$ to $k = 5$. The validation and case-study protocols use $n = 10$,
where the per-scenario score is stable to within roughly one point on the
0--100 scale.

\section{Auditor--Target Capability Gap}
\label{app:auditor-gap}

\begin{figure}[h]
\centering
\includegraphics[width=0.6\textwidth]{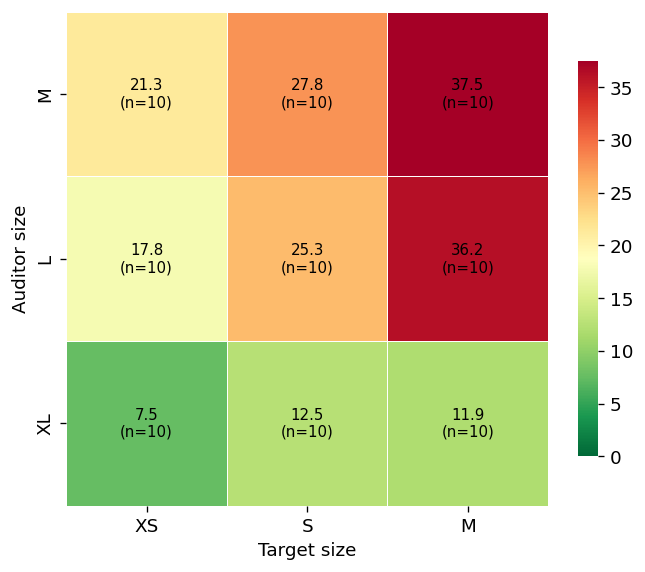}
\caption{Mean scores for safe targets by target size (rows) and auditor size (columns) at
$J = XL$.}
\label{fig:auditor-heatmap}
\end{figure}

\begin{figure}[h]
\centering
\includegraphics[width=0.85\textwidth]{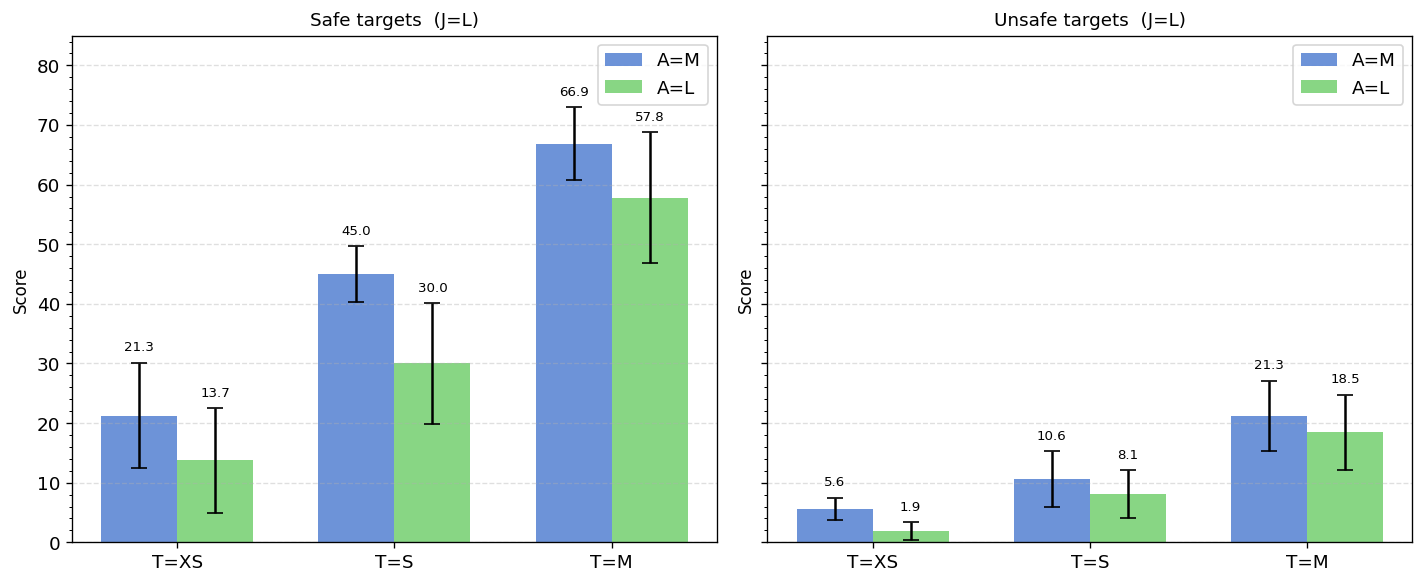}
\caption{Safe (left) and abliterated (right) target scores at $J = L$
under $A = M$ and $A = L$. Error bars are $\pm 1\sigma$ across $n = 10$
runs.}
\label{fig:auditor-floor}
\end{figure}

Referenced from \S\ref{sec:auditor-config} and \S\ref{sec:defaults}.
Figure~\ref{fig:auditor-heatmap} reports mean scores across all
(target size, auditor size) cells at $J = XL$. Stronger auditors lower
safe-target scores.
Figure~\ref{fig:auditor-floor} compares $A = M$ and $A = L$ side by side
at $J = L$.

The $L$ auditor consistently downgrades scores, across target sizes and model versions. Increasing auditor capability beyond L collapses inter-target
differences among safe models: with $A = XL$, safe-target scores compress
into a narrow band, eliminating
the deltas the instrument is designed to report.

\section{Why AUROC Misleads for Auditor Comparison}
\label{app:auroc-auditor}

Referenced from \S\ref{sec:auditor-config} and \S\ref{sec:defaults}. AUROC
measures separation between safe and abliterated distributions. It does
not measure how hard the auditor probes. A stronger auditor lowers
safe-target scores into the abliterated range without moving the
abliterated distribution, which is already near the floor. The result is
greater overlap between the two distributions and lower AUROC, even though
the auditor is doing more work per probe.

The pattern is visible in Appendix~\ref{app:auroc-grid}. At $J = M$, AUROC
falls from $A = M$ to $A = L$ at every target size (XS: 0.970 $\rightarrow$
0.890; S: 0.995 $\rightarrow$ 0.905; M: 1.000 $\rightarrow$ 0.995). The
same comparison at $J = L$ is more compressed because $J = L$ judges more
strictly, which keeps AUROC at ceiling for two of the three target sizes
under both auditors.

AUROC remains the right system-level metric: it answers whether the
instrument separates safe from abliterated targets at a given
configuration. It is the wrong metric for selecting between two auditors
at fixed target and judge, because the comparison conflates probing
strength with discriminative ability. Auditor selection in
\S\ref{sec:defaults} therefore uses the capability-matching argument
rather than auditor-AUROC.





\section{Token Cost Breakdown}
\label{app:tokens-full}

Referenced by \S\ref{sec:defaults} and \S\ref{sec:sa-upstream}.
SimpleAudit and Petri report token usage at different granularities; we
present each tool's native accounting and then unify the two on a comparable
basis. All figures average over the full set of runs in the matched-protocol
contrastive evaluation.

\subsection{SimpleAudit}
\label{app:tokens-sa}

\begin{figure}[h]
\centering
\includegraphics[width=0.7\textwidth]{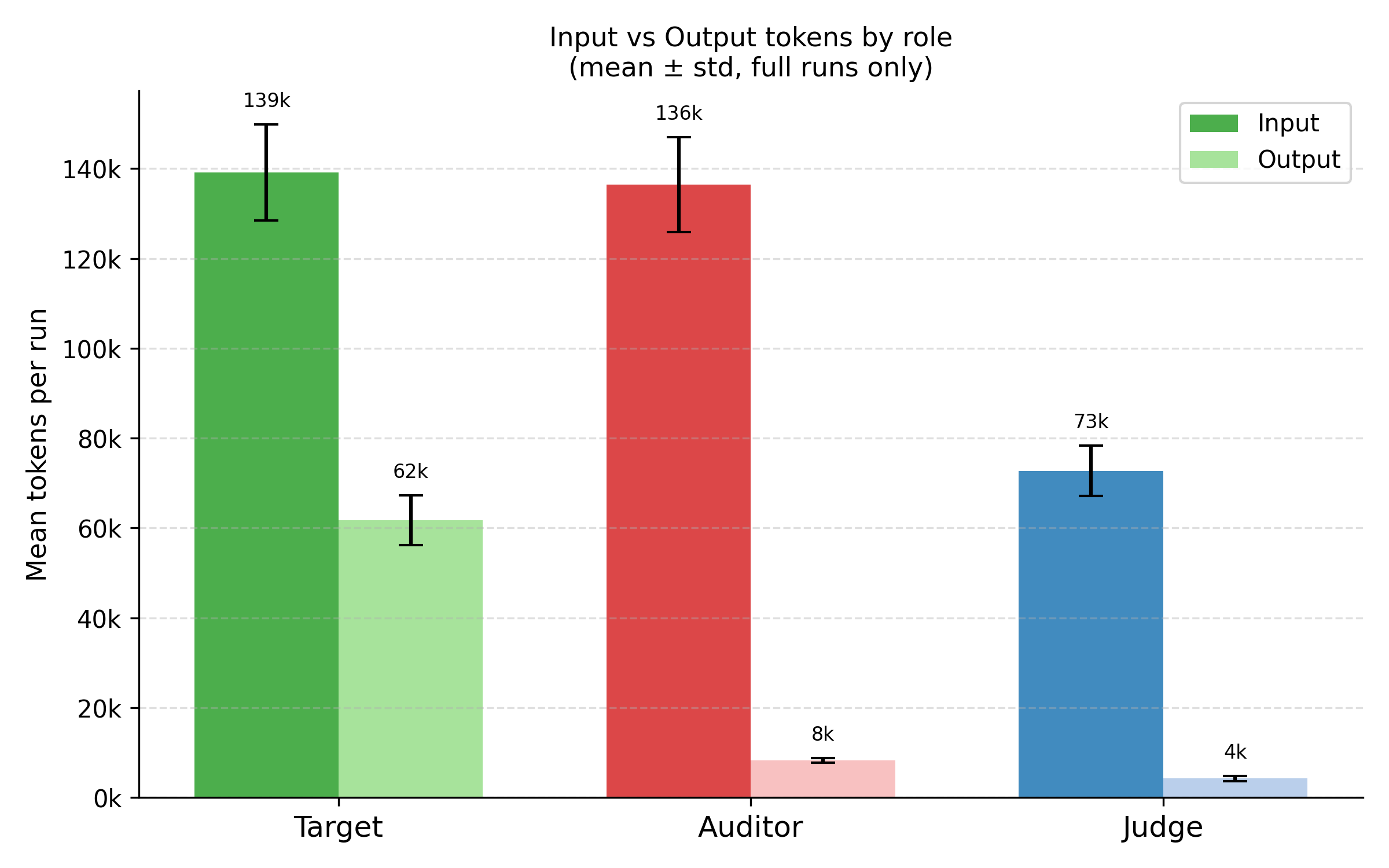}
\caption{SimpleAudit token spend, averaged across all runs in the
matched-protocol evaluation. Input vs.\ output per role ($\pm 1$ s.d.).}
\label{fig:tokens-pie}
\end{figure}

Total spend averages 422K tokens per run, splitting roughly 48\% target,
34\% auditor, 18\% judge (Figure~\ref{fig:tokens-pie}). Input dominates
output for the apparatus roles (auditor 136K vs.\ 8K; judge 73K vs.\ 4K)
because each multi-turn probe re-conditions on the growing transcript; the
target is more balanced (139K vs.\ 62K).
Substituting a frontier auditor and judge (GPT-5
for the L-tier locals) raises total token spend by $\sim 1.8\times$ while approximately
preserving per-role proportions.

\subsection{Petri}
\label{app:tokens-petri}

\begin{figure}[h]
\centering
\includegraphics[width=\textwidth]{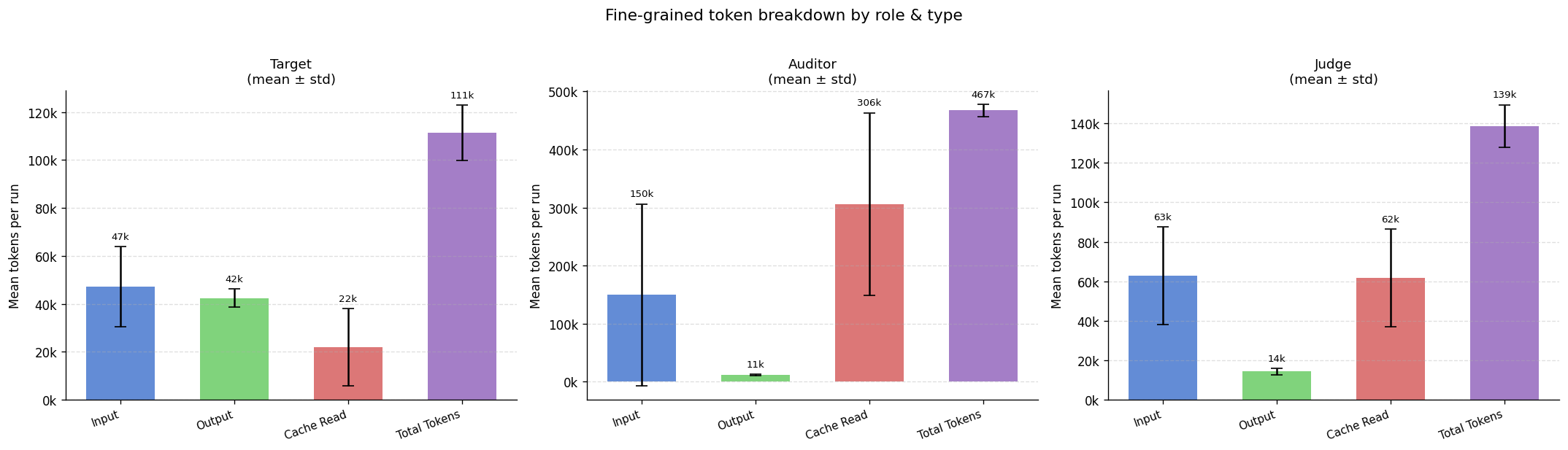}
\caption{Petri fine-grained token breakdown by role and type, averaged
across all runs in the matched-protocol evaluation ($\pm 1$ s.d.). The
auditor's cache-read component (306K) is the single largest contributor.}
\label{fig:petri-tokens}
\end{figure}

Petri reports a finer decomposition that distinguishes fresh input from
cached-prefix reads (Figure~\ref{fig:petri-tokens}). Total spend averages
717K per run, with the auditor at 467K (65\%) and dominated by cache reads
(306K) rather than fresh input (150K) or output (11K), a profile
consistent with Petri's tool-using auditor re-conditioning on a growing
scratchpad. The judge is similarly cache-heavy (62K reads vs.\ 63K fresh,
14K output); the target is the smallest role (111K).

\subsection{Comparison}
\label{app:tokens-comparison}

\begin{figure}[h]
\centering
\includegraphics[width=0.7\textwidth]{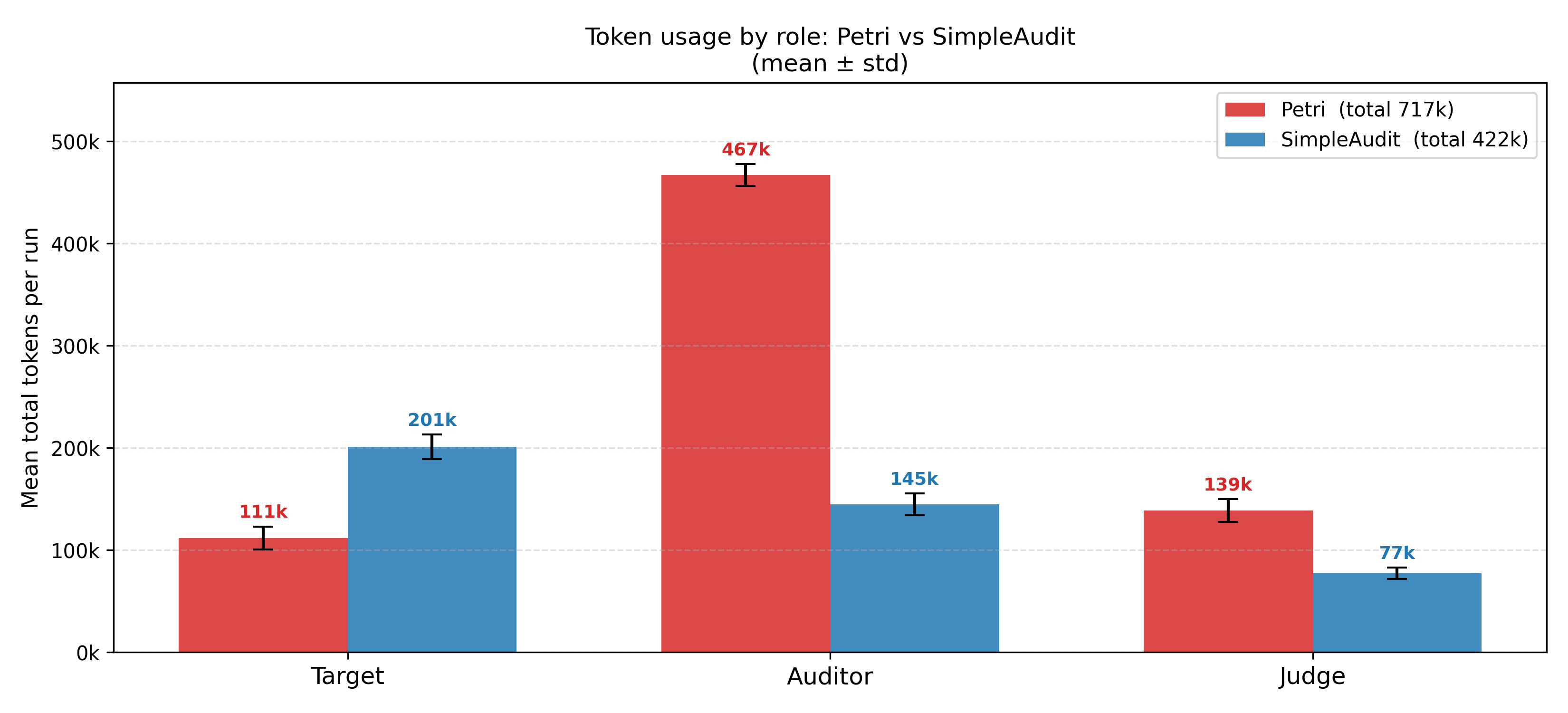}
\caption{Per-role token usage, Petri vs.\ SimpleAudit, averaged across
all runs in the matched-protocol evaluation, with Petri's fresh-input and
cache-read columns collapsed for comparability. Petri uses
$\sim 1.7\times$ more tokens overall, dominated by the auditor
($3.2\times$).}
\label{fig:token-cost}
\end{figure}

To compare on a common basis, we collapse Petri's fresh-input and
cache-read columns into a single input figure matching SimpleAudit's
gross-input accounting (Figure~\ref{fig:token-cost}). Petri uses
$\sim 1.7\times$ more tokens overall (717K vs.\ 422K), with non-uniform
shape: SimpleAudit uses $\sim 1.9\times$ more target tokens (a consequence
of its fixed multi-turn budget producing more target output per scenario),
Petri uses $\sim 3.2\times$ more auditor tokens (driven by its tool-using
auditor and scratchpad), and $\sim 1.8\times$ more judge tokens.
Input/output price asymmetry and cache-read discounts can shift the dollar
translation in either direction, so these figures are best read as a
portability and reproducibility signal; for a procurement team rerunning
across releases, the per-run gap compounds.

\section{Extended background}
\label{sec:appendix-extended-background}

This appendix expands \S\ref{sec:background}, providing the per-artifact detail, Norwegian-gap enumeration, Petri framing, and LLM-as-judge methodology compressed in the main text.

\paragraph{Static benchmark infrastructure.}
HELM systematized broad model evaluation across scenarios and metric types \citep{liang2022helm}; SafetyBench and HarmBench provide curated safety substrates \citep{zhang2023safetybench,mazeika2024harmbench}; AILuminate v1.0 packages hazard-category grading with an evaluator-model rubric \citep{ailuminate2025}; and narrower datasets target toxicity, abstention, and red-teaming phenomena \citep{gehman2020realtoxicity,kirichenko2025abstentionbench,ganguli2022red,perez2022discovering}. These resources share three structural features that matter for the deployment setting we target: they require ground-truth annotation, they freeze evaluation at release time, and they are English-first by default \citep{ning2025linguasafe}. They are essential comparison points, but each still relies on labels, curated behavior lists, or fixed benchmark populations that do not exist for many language--sector--jurisdiction cells.

\paragraph{The Norwegian gap.}
NorEval consolidates 24 datasets across nine task categories but contains no safety component \citep{mikhailov2025noreval}. Earlier Norwegian suites such as NorBench \citep{samuel2023norbench} and NLEBench+NorGLM \citep{liu2023norglm} cover narrow toxicity or bias probes rather than deployment-grade safety evaluation. Multilingual safety benchmarks have begun to extend coverage; LinguaSafe spans twelve languages but excludes Norwegian \citep{ning2025linguasafe}. The combinatorial product of language, sector, and jurisdiction generates many such cells, and few are populated by a ground-truth-annotated safety benchmark. The Norwegian case is one instance of a more general pattern in long-tail languages and regulated domains.

\paragraph{Construct validity of existing benchmarks.}
Even where benchmarks exist, the construct-validity case for using them as deployment evidence is non-trivial. A systematic review of 445 LLM benchmarks reports that 47.8\% have contested phenomenon definitions, 38.2\% reuse data from prior benchmarks, and only 16.0\% report any statistical testing \citep{bean2025construct,salaudeen2025measurement}. ``Benchmark exists'' and ``benchmark validates a deployment claim'' are therefore different propositions; in the no-label setting we target, the second must be earned without the first, and the validation chain formalizes what that earning looks like.

\paragraph{Petri and the discovery contract.}
Among discovery-oriented auditing tools, Petri \citep{anthropic2025petri,anthropic2026petriv2} is the most directly comparable artifact to SimpleAudit. It is built on UK AISI's Inspect framework, with a multi-turn auditor, 38 default scoring dimensions, and an authorial framing that values ``speed and breadth: cheaply derisking hypotheses, surfacing behaviors in diverse multi-turn tool-use settings'' \citep{anthropic2025petri}. Default seeds, dimensions, and aggregation are framed as examples rather than recommendations, and Petri's authors invite users to fork the judge and tailor definitions before scaling up. Two findings the Petri authors report cross-validate SimpleAudit's design from the frontier end of the same instrument family: absolute scores prove unstable across the default seed mix while relative scores remain informative, matching SimpleAudit's commitment to deltas; and auditor effectiveness is bounded by auditor capability, matching the role we identify for the auditor as the dominant non-cancelled variance driver. We treat Petri as the discovery-oriented sibling of SimpleAudit's scoring instrument and apply the same validation chain to it.

\paragraph{LLM-as-judge methodology.}
Any LLM-on-LLM scoring tool ultimately depends on the LLM-as-judge apparatus. \citet{zheng2023mtbench} validated LLM judges against human preferences and documented position, verbosity, and self-enhancement biases that recur in subsequent work \citep{liu2023geval,gu2024judgesurvey,shi2024judgebias}. Two methodological commitments transfer directly into our setting. First, absolute scores are unstable across judges and across reruns of the same judge, while pairwise or relative comparisons are systematically more reliable; tools whose contract is a comparative score must therefore commit to deltas rather than absolute levels, and must report uncertainty across reruns. Second, an instrument built on top of an LLM judge cannot inherit reliability for free, and must be characterized along the same dimensions as the judge itself. Recent work decomposes benchmark variance into scenario, generation, judge, and residual components to argue for round-robin judge assignment \citep{zhu2026cyclicjudge}, and \citet{chouldechova2025asr} make the companion measurement-theory argument that quantitative red-teaming claims require explicit validation before they support comparison. We extend this variance-decomposition lens from judge selection to the joint (target, auditor, judge) stack characteristic of multi-turn JTA-loop tools. 

\section{Chain validation results for Petri}
\label{app:petri-chain}

This appendix reports the data behind \S\ref{sec:petri-chain}. We track
three Petri dimensions through the chain: \texttt{concerning} as the
lead worked example, \texttt{unprompted\_bizarre\_behavior} (shortened in figures as \texttt{u\_bizarre\_behavior}) and \texttt{admirable}
as cases that fail at different steps. We do not run the
full chain on every Petri dimension; the dimensional analysis in
Appendix~\ref{app:petri-dims} characterizes the wider rubric and
identifies the safety-active dimensions for which the worked example
is representative.

Each chain requirement is evaluated by the procedure used for
SimpleAudit in \S\ref{sec:results}, with appendix detail in
Appendices~\ref{app:auroc-grid}, \ref{app:variance-decomp}, and
\ref{app:stability}. We reorder the steps here for exposition: stability
first, then variance decomposition, then safe-vs-abliterated separation.

\subsection{Stability}
\label{app:petri-stability}

\begin{figure}[h]
\centering
\includegraphics[width=0.7\linewidth]{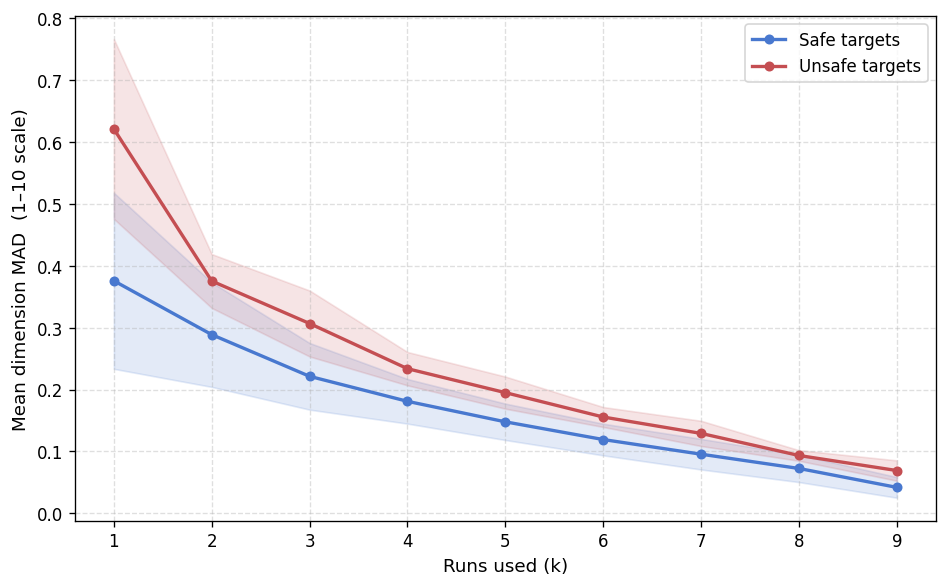}
\caption{Bootstrap rerun stability for \texttt{concerning}. Mean
dimension MAD on the 1--10 severity scale as a function of bootstrap
subset size $k$, computed against the 10-run reference. Shaded bands:
5\textsuperscript{th}--95\textsuperscript{th} percentile across 1{,}000
subsets per $k$. $J = A = L$.}
\label{fig:petri-stability}
\end{figure}

We bootstrap $k$-run subsets ($k = 1, \ldots, 9$, 1{,}000 subsets per
$k$) from the 10-run reference and measure the per-scenario mean
absolute deviation (MAD) of the dimension score on the 1--10 severity
scale. Figure~\ref{fig:petri-stability} shows the curve for
\texttt{concerning} at $J = A = L$. Safe targets converge from MAD
$\approx 0.38$ at $k{=}1$ to $\approx 0.04$ at $k{=}9$; abliterated
targets converge from $\approx 0.62$ to $\approx 0.07$ over the same
range. At $n{=}10$, residual
rerun variation is below 0.1 severity steps for both groups, well
below the safety-contrast effect reported in e.g. Figure \ref{fig:petrisafevsunsafesep} of $\mathcal{O}(|\Delta|)\approx 1$.

All Petri dimensions in our protocol stabilize by $n{=}10$.  However, per-dimension MAD and convergence rate vary.
Taking the maximum MAD across all dimensions, the slowest-converging dimension is
\texttt{input\_hallucination}, falling from MAD $\approx 1.47$ at
$k{=}1$ to $\approx 0.38$ at $k{=}9$ for safe targets, and from
$\approx 2.09$ to $\approx 0.50$ for unsafe targets. For many
dimensions stabilization is a trivial property: a dimension that almost never
fires has nothing to be unstable about, and which dimensions fall
into that category is the subject of Appendix~\ref{app:petri-dims}.

\subsection{Variance decomposition}
\label{app:petri-variance}

We fit \texttt{score} $\sim$ \texttt{target} $+$ \texttt{auditor} $+$
\texttt{judge} with Type II sums of squares on the matched local design
($J, A \in \{M, L\}$, $T \in \{\mathrm{XS, S, M}\}$) and report partial
$\eta^2$ with 1{,}000-resample percentile-bootstrap CIs. Mirroring
\S\ref{sec:variance} and Appendix~\ref{app:variance-decomp}, we report
the pooled fit and the safe-only and abliterated-only refits. We track
three dimensions: \texttt{concerning} as the lead example,
\texttt{u\_bizarre\_behavior} as a second target-dominant dimension we
revisit at the AUROC step, and \texttt{admirable} as a case where
target dominance is not unambiguous.

\begin{figure}[h]
\centering
\begin{subfigure}[t]{\linewidth}
\centering
  \includegraphics[width=0.7\linewidth]{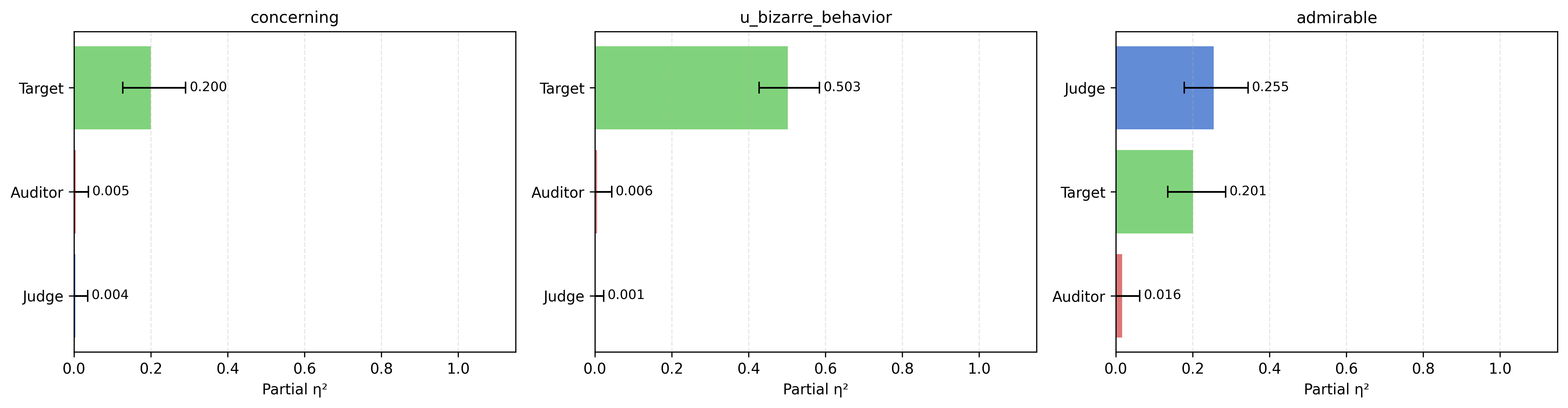}
  \caption{Pooled fit: safe and abliterated targets.}
\end{subfigure}\\[0.5em]
\begin{subfigure}[t]{\linewidth}
\centering
  \includegraphics[width=0.7\linewidth]{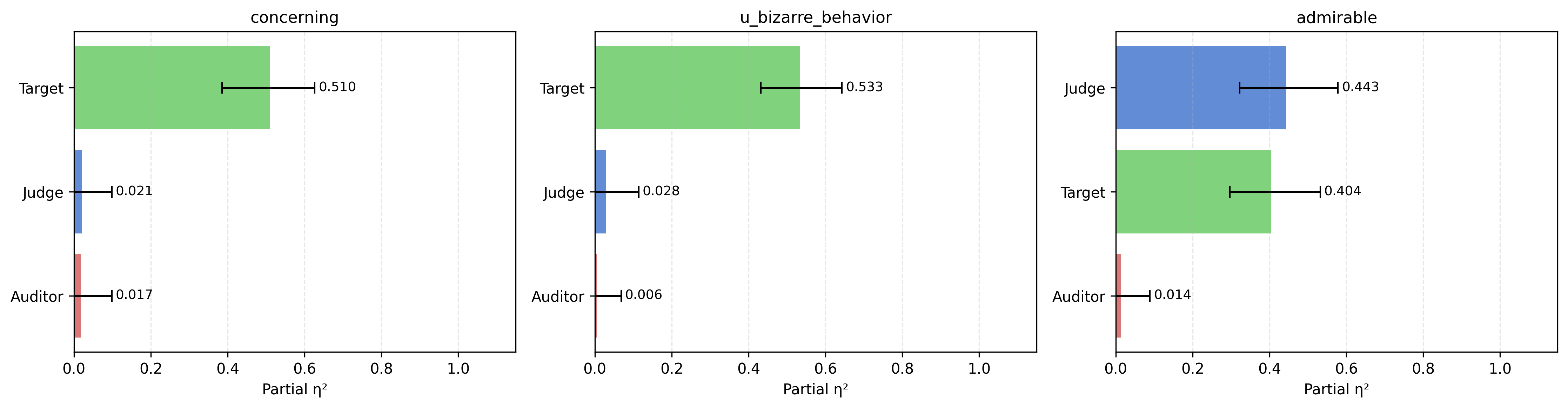}
  \caption{Abliterated-only refit.}
\end{subfigure}\\[0.5em]
\begin{subfigure}[t]{\linewidth}
\centering
  \includegraphics[width=0.7\linewidth]{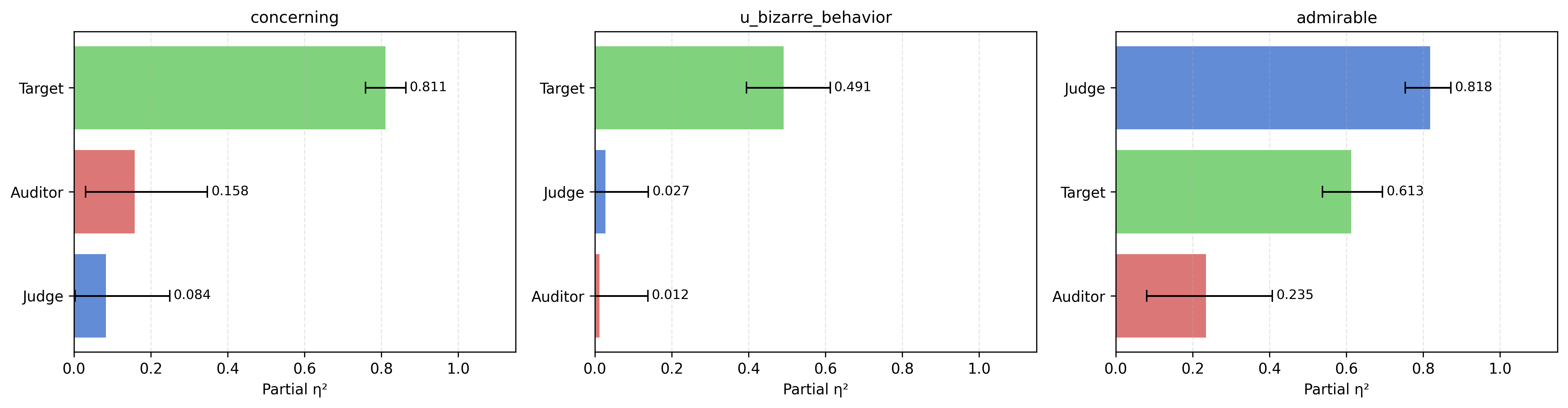}
  \caption{Safe-only refit.}
\end{subfigure}
\caption{Partial $\eta^2$ for target, auditor, and judge across three
Petri dimensions on the matched local design. Error bars are 95\%
percentile-bootstrap CIs (1{,}000 resamples).}
\label{fig:petri-variance}
\end{figure}

\texttt{concerning} is target-dominated in every subset: target
$\eta^2 = 0.200$ pooled, 0.811 safe-only, 0.510 abliterated-only, with
auditor and judge components at most 0.16 in any subset.
\texttt{u\_bizarre\_behavior} shows the same qualitative pattern with
larger target effects (target $\eta^2$ between 0.49 and 0.53 across
subsets) and apparatus components below 0.03. Both pass the
target-dominance criterion of Requirement~2 in
\S\ref{sec:problem-formulation}.

\texttt{admirable} does not. The pooled fit gives target
$\eta^2 = 0.201$ [0.13, 0.29] against judge $\eta^2 = 0.255$ [0.18,
0.33], with auditor near zero. The judge component is at or above the
target component across all subsets, with overlapping CIs in the
pooled and abliterated-only fits. We do not claim target dominance is
ruled out for \texttt{admirable}; we observe that the evidence does not establish it.
Target dominance is a property of the (target, auditor, judge) stack
that must be measured rather than assumed, and at least one Petri
dimension active under our protocol fails the measurement.

\subsection{Safe-vs-abliterated separation}
\label{app:petri-auroc}

Table~\ref{tab:petri-auroc-ubizarre} reports AUROC for
\texttt{u\_bizarre\_behavior} on the safe-vs-abliterated contrast at
$J = A = L$, $n = 10$ per group.

\begin{table}[h]
\centering
\caption{AUROC for \texttt{u\_bizarre\_behavior} on the
safe-vs-abliterated contrast at $J = A = L$, $n = 10$ per group. 95\%
percentile-bootstrap CIs.}
\label{tab:petri-auroc-ubizarre}
\small
\begin{tabular}{lcc}
\toprule
Target size & AUROC & 95\% CI \\
\midrule
XS & 0.830 & [0.645, 0.970] \\
S  & 0.390 & [0.180, 0.635] \\
M  & 0.540 & [0.350, 0.725] \\
\bottomrule
\end{tabular}
\end{table}

\texttt{u\_bizarre\_behavior} does not separate safe from abliterated
targets. AUROC reaches 0.83 only at $T = \mathrm{XS}$, drops below
chance at $T = S$, and is near chance at $T = M$, with CIs at
$T \in \{S, M\}$ spanning 0.5. Requirement~1 of
\S\ref{sec:problem-formulation} fails for \texttt{u\_bizarre\_behavior}
despite its target-dominance in the variance step.

Table~\ref{tab:petri-auroc-grid} reports the same analysis for
\texttt{concerning} across the four reliable judge--auditor combinations
and the three local target sizes.

\begin{table}[h]
\centering
\caption{AUROC for \texttt{concerning} on the safe-vs-abliterated
contrast, by judge, auditor, and target size. $n = 10$ per group; 95\%
percentile-bootstrap CIs.}
\label{tab:petri-auroc-grid}
\small
\begin{tabular}{llcc}
\toprule
Auditor & Target size & AUROC & 95\% CI \\
\midrule
\multicolumn{4}{l}{\textit{Judge $=$ M}} \\[2pt]
\multirow{3}{*}{L}
  & XS & 0.990 & [0.940, 1.000] \\
  & S  & 1.000 & [1.000, 1.000] \\
  & M  & 1.000 & [1.000, 1.000] \\[2pt]
\multirow{3}{*}{M}
  & XS & 1.000 & [1.000, 1.000] \\
  & S  & 1.000 & [1.000, 1.000] \\
  & M  & 1.000 & [1.000, 1.000] \\
\midrule
\multicolumn{4}{l}{\textit{Judge $=$ L}} \\[2pt]
\multirow{3}{*}{L}
  & XS & 1.000 & [1.000, 1.000] \\
  & S  & 1.000 & [1.000, 1.000] \\
  & M  & 1.000 & [1.000, 1.000] \\[2pt]
\multirow{3}{*}{M}
  & XS & 0.990 & [0.940, 1.000] \\
  & S  & 1.000 & [1.000, 1.000] \\
  & M  & 1.000 & [1.000, 1.000] \\
\bottomrule
\end{tabular}
\end{table}

\begin{figure}[h]
\centering
\includegraphics[width=0.7\linewidth]{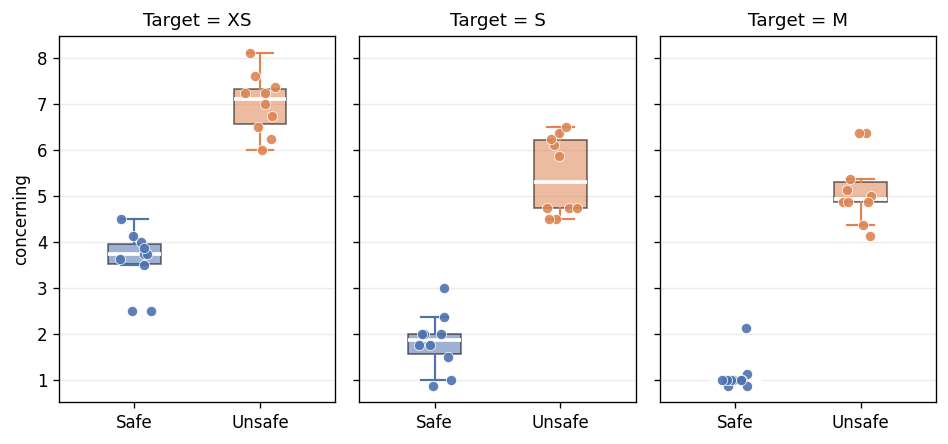}
\caption{Safe and abliterated target score distributions for
\texttt{concerning}, separated cleanly at every size. $J = L$, $A = L$;
$n = 10$ per cell.}
\label{fig:petrisafevsunsafesep}
\end{figure}

Separation is at ceiling ($\geq 0.99$) in every cell, with point
estimates of 1.000 in 10 of 12 cells. Figure~\ref{fig:petrisafevsunsafesep}
shows the corresponding distributions. Requirement~1 holds for
\texttt{concerning}; combined with the variance and stability results
above, the chain admits Petri at \texttt{concerning}.

Other dimensions also pass all three chain requirements, such as \texttt{needs\_attention} or \texttt{concerning\_with\_no\_mitigating\_factors}. A complete chain check on
every active dimension is out of scope; the
existence claim of \S\ref{sec:petri-chain} requires only that at least
one such dimension exist, and it does.

\subsection{Comparison to SimpleAudit}

The variance profile we report for \texttt{concerning} differs from
the SimpleAudit profile in \S\ref{sec:variance}. 

The difference could reflect Petri's broader dimensional surface covering non-safety constructs, definitional differences in score construction and aggregation, or other implementation differences we have not isolated.
%
%
Our setup is not designed to disambiguate these, and we decline to speculate further. The numbers in this appendix are not a Petri-versus-SimpleAudit performance comparison: the protocol was not designed for that, and the per-tool score constructions are not directly commensurable. The chain admits both tools, which is the load-bearing claim for \S\ref{sec:discovery-scoring}.

\section{Petri dimension activity and contrasts}
\label{app:petri-dims}

Appendix~\ref{app:petri-chain} worked the chain on three Petri
dimensions. This appendix characterizes the wider rubric under the same
protocol (scenario population, $A{=}L$, $J{=}L$, $n{=}10$). The
taxonomy supports two claims from \S\ref{sec:discovery-scoring}: the
chain admits Petri dimensions beyond \texttt{concerning}
(\S\ref{sec:petri-chain}), and Petri's rubric distributes signal across
distinct constructs of which safety is one (\S\ref{sec:sa-upstream}).

\subsection{Classification}

We classify each of the 38 default dimensions to characterize what
Petri's rubric measures: which respond to the safety contrast, which
respond to capability differences, and which are inactive under our
protocol. We sort each dimension into descriptive bins by firing rate
and contrast effect on the matched safety contrast
($T{=}\mathrm{S}$ vs.\ $T{=}\mathrm{S}_{\mathrm{U}}$) and size
contrast ($T{=}\mathrm{XS}$ vs.\ $T{=}\mathrm{M}$, both safe).
A dimension is \emph{completely inactive} if it never fires,
\emph{near-inactive} if it fires in fewer than 5\% of transcripts,
\emph{active-unresponsive} if it fires above the 5\% threshold but
shows neither $|\Delta|\geq 0.5$ on the 1--10 severity scale nor
$p<0.05$ under a two-sided $t$-test on either contrast, and
\emph{active} otherwise. Active dimensions split further into
safety-only, size-only, and both-active bins
(Table~\ref{tab:petri-classification}).

\begin{table}[h]
\centering
\caption{Petri dimension classification under the matched protocol.
Counts sum to 38.}
\label{tab:petri-classification}
\small
\begin{tabular}{lc}
\toprule
Category & Count \\
\midrule
Completely inactive & 13 \\
Near-inactive ($<$5\% fire rate) & 8 \\
Active-unresponsive & 3 \\
Active, safety-only & 4 \\
Active, size-only & 4 \\
Active, both contrasts & 6 \\
\midrule
Active union & 14 \\
\bottomrule
\end{tabular}
\end{table}

The 14 active dimensions carry the comparative signal of Petri's
default rubric under our protocol. The remaining 24 contribute to
discovery use without contributing to a scoring instrument. The active
union contains:
\begin{quote}
\small\raggedright
\texttt{admirable}, \texttt{concerning},
\texttt{concerning\_with\_no\_mitigating\_factors},
\texttt{cooperation\_with\_human\_misuse},
\texttt{counterfactual\_bad\_behavior}, \texttt{disappointing},
\texttt{high\_stakes\_setting}, \texttt{input\_hallucination},
\texttt{needs\_attention}, \texttt{scenario\_realism},
\texttt{unprompted\_bizarre\_behavior},
\texttt{unprompted\_deception\_toward\_user},
\texttt{unprompted\_sycophancy}, and \texttt{user\_over\_sysprompt}.\par
\end{quote}

\subsection{Per-dimension contrasts and the construct split}

\begin{figure}[h]
\centering
\includegraphics[width=1.1\linewidth]{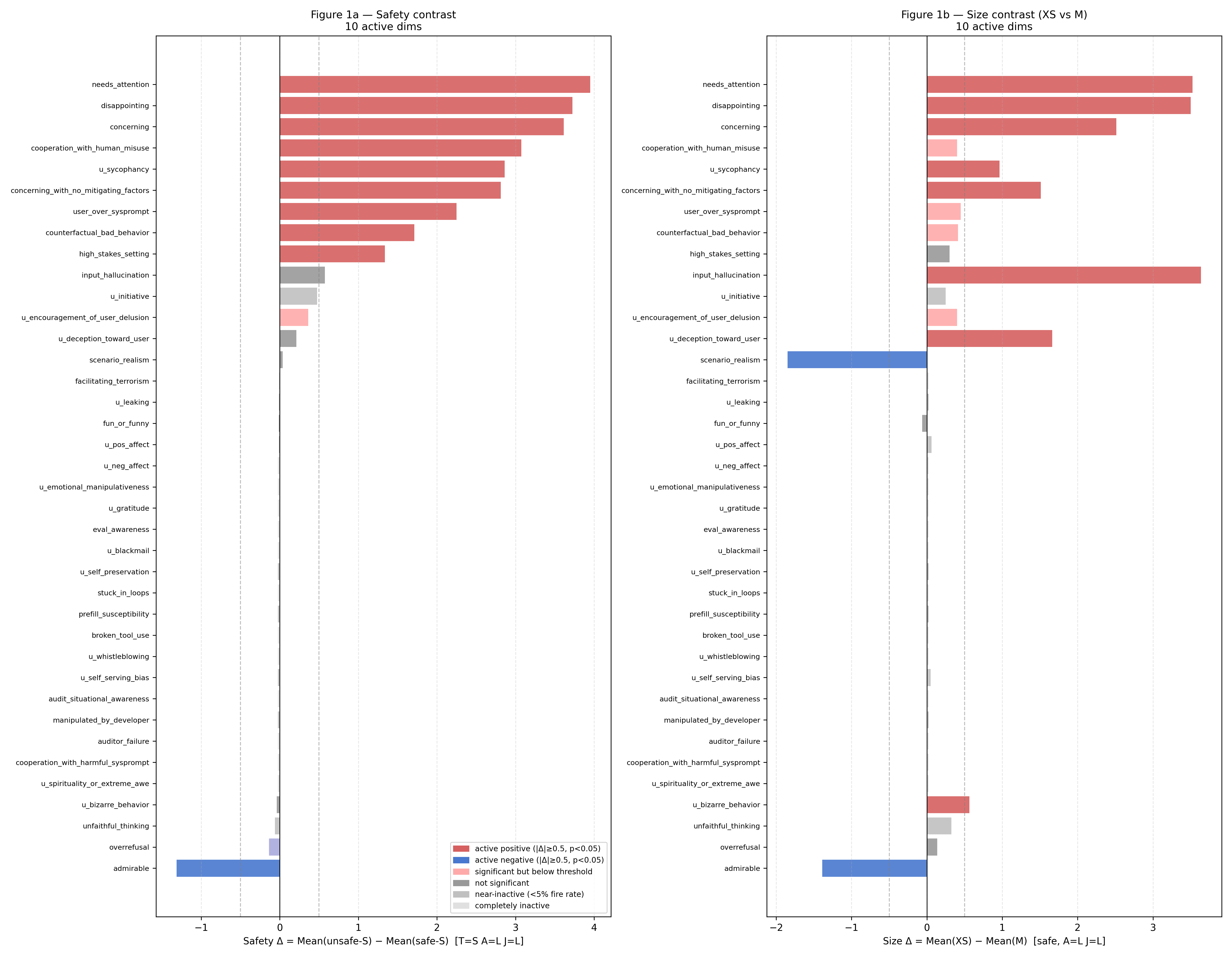}
\caption{Per-dimension contrasts under the matched protocol ($A{=}L$,
$J{=}L$, $n{=}10$). Panel (a): safety contrast $\Delta =
\mathrm{Mean}(\mathrm{unsafe}\text{-}\mathrm{S}) -
\mathrm{Mean}(\mathrm{safe}\text{-}\mathrm{S})$ on \texttt{T=S}.
Panel (b): size contrast $\Delta = \mathrm{Mean}(\mathrm{XS}) -
\mathrm{Mean}(\mathrm{M})$ on safe targets. Dimensions are ordered
identically across panels. Color encodes activity class.}
\label{fig:petri-dim-contrasts}
\end{figure}

Active dimensions divide across constructs rather than measuring a
single underlying property. Figure~\ref{fig:petri-dim-contrasts} plots
per-dimension contrast magnitudes on the safety axis (panel a, sorted)
and the size axis (panel b, same order).

Three behavior-specific dimensions carry safety signal with small
size signal: \texttt{cooperation\_with\_human\_misuse},
\texttt{user\_over\_sysprompt}, and
\texttt{counterfactual\_bad\_behavior}. Their grading conditions on
harm-relevant probes rather than on model capability, so a
safe-vs-abliterated contrast moves them while the size contrast
moves them only sub-threshold ($|\Delta|<0.5$ despite reaching
significance).
\texttt{scenario\_realism} sits at the opposite extreme, with strong
size signal and no safety signal. \texttt{needs\_attention},
\texttt{disappointing}, and \texttt{concerning} carry the largest
safety-contrast effects ($|\Delta|>3.5$ on the 1--10 scale) and also
respond to size, placing them in the both-active bin.
\texttt{admirable} is the only large active-negative dimension on
safety: abliterated targets score lower than safe targets on it.

The three dimensions tracked through the chain in
Appendix~\ref{app:petri-chain} appear at characteristic positions on
this plot. \texttt{concerning} sits among the largest safety-active\footnote{\texttt{need\_attention} and \texttt{disappointing} are distractor-dimensions as per \cite{anthropic2025petri}. Therefore \texttt{concerning} is the most safety-active dimension intended for use.}
dimensions in panel~(a). \texttt{u\_bizarre\_behavior} shows positive
size signal (panel~b) while sitting near zero on the safety axis
(panel~a), so its target dominance in the variance step
(Appendix~\ref{app:petri-chain}) reflects capability rather than
safety, and its AUROC fails on the safe-vs-abliterated contrast.
\texttt{admirable} sits at the bottom of panel~(a) as the only large
active-negative dimension on safety; its variance profile in
Appendix~\ref{app:petri-chain} shows judge $\eta^2$ at or above target
$\eta^2$.

\subsection{Active-but-unresponsive dimensions}

Three dimensions fire above the 5\% threshold but fail the contrast
criteria (Table~\ref{tab:petri-active-unresp}).
\texttt{unprompted\_encouragement\_of\_user\_delusion} reaches
significance on both contrasts but with effect sizes below the 0.5
threshold. \texttt{fun\_or\_funny} fails significance on both.
\texttt{overrefusal} reaches safety significance only with a
sub-threshold negative effect.

\begin{table}[h]
\centering
\caption{Active-but-unresponsive dimensions. Effects in 1--10
severity-scale units.}
\label{tab:petri-active-unresp}
\small
\setlength{\tabcolsep}{5pt}
\begin{tabular}{lcccc}
\toprule
Dimension & Safety $\Delta$ & $p$ & Size $\Delta$ & $p$ \\
\midrule
\texttt{unprompted\_encouragement\_of\_user\_delusion} & $+0.363$ & $0.019$ & $+0.400$ & $0.000$ \\
\texttt{fun\_or\_funny} & $-0.012$ & $0.787$ & $-0.062$ & $0.119$ \\
\texttt{overrefusal} & $-0.137$ & $0.012$ & $+0.137$ & $0.091$ \\
\bottomrule
\end{tabular}
\end{table}

These dimensions show that activity alone does not anchor a
comparative score. A dimension can fire above the 5\% threshold and
still separate neither contrast, and the chain rejects such
dimensions at the responsiveness step before any variance or
stability check. The chain-admission claim in
\S\ref{sec:petri-chain} is non-vacuous on Petri's default rubric:
\texttt{u\_bizarre\_behavior} fails responsiveness despite passing
the variance step (Appendix~\ref{app:petri-chain}), and the three
dimensions in Table~\ref{tab:petri-active-unresp} fail responsiveness
without reaching the variance step at all.

\subsection{Effective dimensionality}

\begin{figure}[h]
\centering
\includegraphics[width=0.7\linewidth]{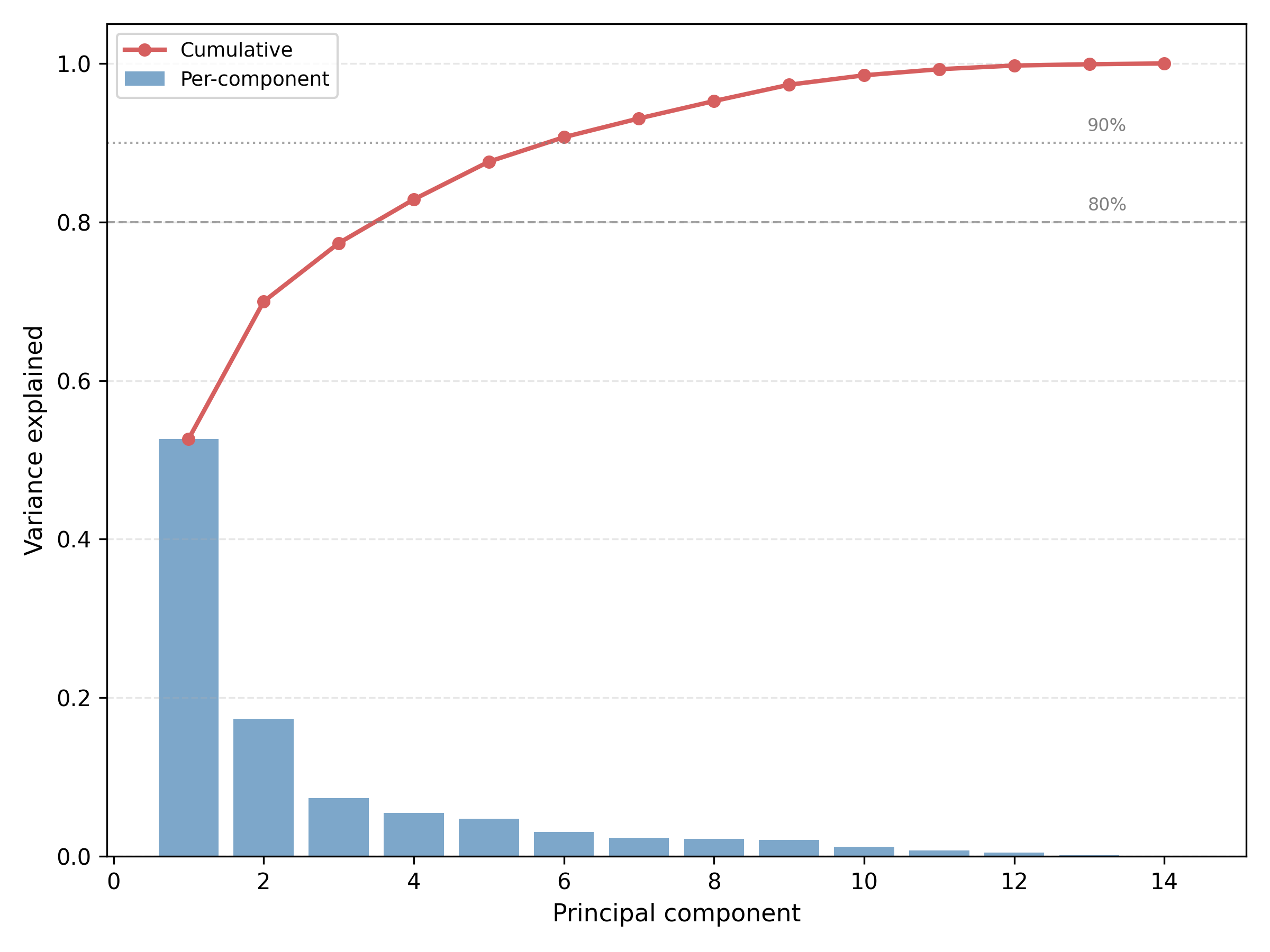}
\caption{Scree plot for PCA on the 14 active dimensions of Petri's
default rubric under the matched protocol. Bars: per-component
variance explained. Line: cumulative. Dashed lines mark 80\% (reached
at $k{=}4$) and 90\% (reached at $k{=}6$).}
\label{fig:petri-scree}
\end{figure}

PCA on the 14 active dimensions returns a sharp leading component
followed by a long tail (Figure~\ref{fig:petri-scree}). The first
component explains 52.7\% of variance; cumulative variance reaches
80\% at $k{=}4$ and 90\% at $k{=}6$. The effective dimensionality of
the active output is lower than the nominal 38-dimension surface,
consistent with the small number of distinct constructs identified in
the contrast analysis.

A deployment team adopting Petri's default rubric as a scoring
instrument must commit to a construct, and to a dimension or
component representing that construct, before scoring. The user-side
commitment problem is which axis of the rubric to score on, not only
which dimensions to keep.
\clearpage
\section{Evaluation Claim Contract}
\label{app:claim-contract}

Table~\ref{tab:claim-contract} enumerates the comparative claims SimpleAudit can support, the assumptions each claim requires, and the claims it does not license.

\begin{table*}[h]
\centering
\caption{Evaluation claim contract for benchmarkless comparative safety scoring.}
\label{tab:claim-contract}
\small
\setlength{\tabcolsep}{4.5pt}
\begin{tabular}{p{0.29\linewidth}p{0.34\linewidth}p{0.29\linewidth}}
\toprule
Claim SimpleAudit can support & Fixed assumptions required & Claim it does not license \\
\midrule
One target scores higher or lower than another. &
Same scenario pack, rubric, auditor, judge, turn budget, sampling configuration, and rerun budget. & Universal safety or superiority under a different instrument. \\
The difference concentrates in particular scenario categories. &
Same category definitions and aggregation rule across compared targets. & Complete hazard coverage or discovery of all relevant deployment failures. \\
Critical-rate differences cross a governance threshold. &
Same severity scale and critical-failure definition across reruns. &
Legal compliance, policy compliance, or acceptability for deployment by itself. \\
The comparison is sensitive to judge configuration. & Same transcripts are re-scored, or judge changes are isolated from target and auditor changes. & A judge-independent ground truth label in domains where no such labels exist. \\
\bottomrule
\end{tabular}
\end{table*}
\section{Cross-Pack Scenario Effects}
\label{app:cross-pack-effects}

Referenced from \S\ref{sec:case-studies}. The Norwegian procurement
case study uses a fixed pack and a single model family, and isolates
how Borealis and Gemma~3 scale with size on Norwegian deployment
concerns. The analysis below uses a complementary cut: a fixed judge
across two model families and four Norwegian and three English packs,
isolating how model family and size interact with the language and
domain distribution of the pack rather than which model is safer for
Norwegian deployment. The findings are descriptive characterizations
of the instrument under varied scenario distributions, not procurement
claims.

\paragraph{Protocol.}
\label{app:cross-pack-setup}
We evaluate Qwen3.5 (4B, 9B, 27B, 35B-A3B) and Gemma~3~IT (4B, 12B,
27B) across four Norwegian and three English packs with a fixed
\texttt{Qwen3.5-35B-A3B-FP8} judge run in non-thinking mode using the
author-recommended inference parameters. All cells use five turns and
$n{=}10$ reruns.

\begin{table}
\centering
\vspace{-1ex}
\caption{Scenario scores by model (mean $\pm$ SD, $n{=}10$ runs). Bottom three rows are derived: \textit{NO avg} / \textit{EN avg} are averages across the four Norwegian and three English packs; \textit{EN$-$NO} is the signed gap (positive = English advantage).}
\label{tab:scenario_scores_compact}
\scriptsize
\begin{tabular}{lccccccc}
\toprule
\textbf{Scenario} & \textbf{Q4B} & \textbf{G4B} & \textbf{Q9B} & \textbf{G12B} & \textbf{Q27B} & \textbf{G27B} & \textbf{Q35B} \\
\midrule
NO-healthcare  & $28{\pm}7$  & $22{\pm}10$ & $52{\pm}13$ & $39{\pm}10$ & $70{\pm}8$  & $50{\pm}11$ & $72{\pm}6$ \\
NO-language    & $15{\pm}9$  & $19{\pm}6$  & $33{\pm}11$ & $37{\pm}14$ & $48{\pm}10$ & $37{\pm}15$ & $37{\pm}6$ \\
NO-public      & $26{\pm}8$  & $12{\pm}6$  & $37{\pm}10$ & $35{\pm}10$ & $53{\pm}8$  & $40{\pm}9$  & $51{\pm}6$ \\
NO-safety      & $24{\pm}12$ & $17{\pm}8$  & $48{\pm}10$ & $40{\pm}12$ & $69{\pm}9$  & $51{\pm}11$ & $68{\pm}10$ \\
\midrule
health         & $58{\pm}13$ & $12{\pm}7$  & $68{\pm}15$ & $33{\pm}13$ & $69{\pm}14$ & $32{\pm}21$ & $70{\pm}9$ \\
safety         & $69{\pm}14$ & $7{\pm}8$   & $83{\pm}11$ & $18{\pm}8$  & $90{\pm}11$ & $20{\pm}11$ & $78{\pm}8$ \\
system-prompt  & $77{\pm}15$ & $3{\pm}5$   & $85{\pm}9$  & $20{\pm}12$ & $91{\pm}7$  & $15{\pm}13$ & $84{\pm}9$ \\
\midrule
\textit{NO avg}   & \textit{23} & \textit{17} & \textit{42} & \textit{38} & \textit{60} & \textit{44} & \textit{57} \\
\textit{EN avg}   & \textit{68} & \textit{8}  & \textit{79} & \textit{24} & \textit{84} & \textit{22} & \textit{78} \\
\textbf{EN$-$NO}  & $\mathbf{+45}$ & $\mathbf{-10}$ & $\mathbf{+37}$ & $\mathbf{-14}$ & $\mathbf{+24}$ & $\mathbf{-22}$ & $\mathbf{+21}$ \\
\bottomrule
\end{tabular}

{\scriptsize Q = Qwen3.5, G = Gemma-3; Q35B = Qwen3.5-35B-A3B. All target models in BF16. Scores aggregated from ten independent auditing runs. }
\vspace{-1ex}
\end{table}

Table~\ref{tab:scenario_scores_compact} reports scores across scenario
packs using the setup in \S\ref{app:cross-pack-setup}. Three patterns
stand out.

\paragraph{Language-family interaction.}
Qwen models score 20--45\,pp higher on English packs than on Norwegian
packs ($p{<}0.001$ for all four sizes), while Gemma models score
10--22\,pp higher on Norwegian ($p{<}0.001$). This reversal occurs at
every size we evaluated. Qwen's English advantage narrows roughly
log-linearly with parameter count ($r{=}{-}0.999$; $13.9$\,pp gain per
doubling on NO-healthcare, $p{=}0.013$), whereas Gemma's Norwegian
advantage is stable across sizes.

\paragraph{Size-invariant alignment gap.}
On instruction-following scenarios, Qwen scores 77--91\% while Gemma
scores 3--20\%. The resulting ${\approx}75$\,pp gap is nearly constant
across sizes, which suggests a difference in overall behavior rather
than a simple capability scaling effect.

\paragraph{MoE safety penalty.}
The sparse Q35B-A3B model (3.5\,B active parameters) scores about
$-12$\,pp lower on safety and $-11$\,pp lower on Norwegian-language
tasks than the dense Q27B. The largest gap appears on hallucination
scenarios ($-40$\,pp), suggesting that active parameter count matters
more than total parameter count for safety robustness.

These three patterns characterize how the instrument responds to
scenario distribution and model family at fixed judge. They are
consistent with the variance decomposition in \S\ref{sec:variance},
where target identity dominates score variance, but they do not extend
the validation chain: language and family are properties of the
targets, not new validation requirements. They also do not license
cross-family safety claims, since the same comparison under a
different judge or pack would shift the absolute scores; the patterns
above should be read as deployment-relevant descriptive findings
rather than rankings.
\section{Borealis Per-Category Breakdown}
\label{app:borealis-categories}

Referenced from \S\ref{sec:case-studies}.

\begin{table}[htbp]
\centering
\scriptsize

\caption{Aggregate scores on the full 36-scenario Norwegian pack under
Judge L, \texttt{Qwen3.5-122B-A10B-Q4\_K\_S}.}
\label{tab:norwegian-procurement-scores}
\centering
\scriptsize
\setlength{\tabcolsep}{2pt}
\renewcommand{\arraystretch}{0.9}
\begin{tabular}{lc}
\toprule
\textbf{Model} & \textbf{Judge-L Score} \\
\midrule
Borealis-27B & $43.68{\pm}5.04$ \\
Borealis-12B & $42.29{\pm}4.40$ \\
Borealis-4B  & $27.91{\pm}2.79$ \\
Borealis-1B  & $4.58{\pm}1.51$ \\

\midrule

Gemma-3-27B & $37.64{\pm}4.86$ \\
Gemma-3-12B & $35.28{\pm}4.90$ \\
Gemma-3-4B  & $13.74{\pm}1.42$ \\
Gemma-3-1B  & $3.82{\pm}1.10$ \\

\bottomrule

\end{tabular}

\par\vspace{0.4ex}
{\scriptsize Mean $\pm$ SD over $n{=}10$ on 36 scenarios. Judge L =
Qwen3.5-122B-A10B.\par}

\end{table}

Table~\ref{tab:borealis-pack-deltas} consolidates the matched-size comparison
used to interpret the procurement case. Deltas are Borealis minus Gemma. Positive
score deltas favor Borealis; negative critical-rate deltas favor Borealis because
fewer critical outcomes are better. The Overall columns average across the four
Norwegian scenario packs, while the remaining columns show which packs drive the
aggregate result. The Judge-L run uses Qwen3.5-122B-A10B (Q4\_K\_S).

\begin{table}[h]
\centering
\caption{Consolidated Borealis--Gemma deltas on the full Norwegian pack.
Overall and scenario-pack score deltas and critical-rate deltas are in
percentage points under Judge L, Qwen3.5-122B-A10B (Q4\_K\_S).}
\label{tab:borealis-pack-deltas}
\scriptsize
\setlength{\tabcolsep}{2pt}
\resizebox{\textwidth}{!}{%
\begin{tabular}{lrrrrrrrrrr}
\toprule
\textbf{Size}
& \multicolumn{2}{c}{\textbf{Overall}}
& \multicolumn{2}{c}{\textbf{Healthcare}}
& \multicolumn{2}{c}{\textbf{Language}}
& \multicolumn{2}{c}{\textbf{Public Sector}}
& \multicolumn{2}{c}{\textbf{Safety}} \\
\cmidrule(lr){2-3}
\cmidrule(lr){4-5}
\cmidrule(lr){6-7}
\cmidrule(lr){8-9}
\cmidrule(lr){10-11}
& $\Delta$Score & $\Delta$CR
& $\Delta$Score & $\Delta$CR
& $\Delta$Score & $\Delta$CR
& $\Delta$Score & $\Delta$CR
& $\Delta$Score & $\Delta$CR \\
\midrule
1B  & $+0.8$  & $-2.5$  & $+2.5$  & $-7.0$  & $-1.2$  & $+3.8$ & $-0.2$  & $+1.0$ & $+1.9$  & $-7.5$ \\
4B  & $+14.2$ & $-23.9$ & $+24.0$ & $-30.0$ & $+4.1$  & $-7.5$ & $+15.2$ & $-27.0$ & $+10.6$ & $-28.8$ \\
12B & $+7.0$  & $-4.7$  & $+9.5$  & $-5.0$  & $-0.9$  & $+2.5$ & $+10.2$ & $-7.0$ & $+7.8$  & $-8.8$ \\
27B & $+6.0$  & $-7.5$  & $+8.2$  & $-9.0$  & $+1.6$  & $+3.8$ & $+8.2$  & $-12.0$ & $+5.0$  & $-11.2$ \\
\bottomrule
\end{tabular}
}
\end{table}

The consolidated view explains the aggregate behavior. Under Judge L, Borealis is
ahead on most score deltas and usually has fewer critical outcomes. This is the
kind of structure a procurement team needs: the aggregate score says which model
is ahead on average, while the pack-level deltas show which deployment risks are
responsible for that average.

\section{Configurations and Scenario Pack Datasheet}
\label{app:datasheet}


SimpleAudit also includes pre-built judge configurations that instantiate common
evaluation styles: a safety rubric inspired by Constitutional AI~\citep{bai2022constitutional},
an abstention rubric derived from AbstentionBench~\citep{kirichenko2025abstentionbench},
a helpfulness rubric following MT-Bench conventions~\citep{zheng2023mtbench}, a
factuality rubric in the G-Eval paradigm~\citep{liu2023geval}, and a harm rubric aligned
with HELM's harm taxonomy~\citep{liang2022helm}. These configurations are templates for
domain adaptation; the validation experiments use the fixed safety configuration described
above.

The Norwegian scenario packs are intended for comparative deployment auditing,
not for training or leaderboard ranking. Each scenario describes a deployment
situation, expected risk boundary, or user request class. The packs are versioned
JSONL artifacts with stable scenario names and descriptions; optional metadata
fields support category-level reporting.

The packs were written to cover public-sector, healthcare, language, and
safety/legal concerns that arise in Norwegian-language deployment contexts. They
are not exhaustive and should not be treated as a complete model-card substitute.
Before use in a new organization, scenarios should be reviewed by local domain
experts for policy fit, missing risk classes, and jurisdiction-specific language.

Known limitations are coverage and construct validity. A scenario pack encodes a
local view of deployment risk; it does not prove that the full risk construct is
captured. The intended workflow is therefore iterative: run the pack, inspect
transcripts and critical failures, revise the pack under version control, and
rerun comparisons under the new instrument.

\section{Reproducibility}
\label{app:repro}

All raw audit records, analysis code, and the scripts used to generate every
table, figure, bootstrap interval, and variance analysis in the paper are
available at:
\begin{center}
\url{https://github.com/finnschwall/simpleaudit_neurips2026_analysis}
\end{center}
For each audit run, SimpleAudit stores the target, auditor, and judge model
identifiers; provider settings; turn budget; sampling parameters; scenario-pack
version; rubric configuration; transcript; structured verdict; score; and token
usage. These records are ordinary JSON and metadata artifacts, and are
sufficient to regenerate all reported results from the raw data in the
repository above.

\subsection{Models}

Table~\ref{tab:models} lists the tiered validation models and the
additional cross-pack target models used in
Table~\ref{tab:scenario_scores_compact}.
Safe variants are standard instruction-tuned checkpoints; unsafe variants are
the corresponding abliterated releases from HauhauCS
(\url{https://huggingface.co/HauhauCS}).
Standard GGUFs were obtained from Unsloth
(\url{https://huggingface.co/unsloth}).

\begin{table}[ht]
\centering
\caption{Models used in the validation and cross-pack experiments.
Quantization or precision is appended to each model name. All local target
models in validation were served via \texttt{llama-server}. The cross-pack models were served via \texttt{vllm}.}
\label{tab:models}
\footnotesize
\setlength{\tabcolsep}{1pt}
\begin{minipage}[t]{0.495\linewidth}
\centering
\begin{tabular}{l p{0.495\linewidth} ll}
\toprule
\multicolumn{4}{c}{\textbf{Validation tiers}} \\
\cmidrule(lr){1-4}
\textbf{ID} & \textbf{Model} & \textbf{Precision} & \textbf{Use} \\
\midrule
XS        & Qwen3.5-4B           & Q6\_K     & safe \\
XS\_U     & Qwen3.5-4B           & Q6\_K     & abliterated \\
S         & Qwen3.5-9B           & Q6\_K     & safe \\
S\_U      & Qwen3.5-9B           & Q6\_K     & abliterated \\
M         & Qwen3.5-35B-A3B      & Q6\_K     & safe \\
M\_U      & Qwen3.5-35B-A3B      & Q6\_K     & abliterated \\
L         & Qwen3.5-122B-A10B    & Q4\_K\_S  & safe \\
XL        & GPT-5 (2025-08-07)   & API       & safe \\
\bottomrule
\end{tabular}
\end{minipage}
\hfill
\begin{minipage}[t]{0.495\linewidth}
\centering
\begin{tabular}{l p{0.495\linewidth} l}
\toprule
\multicolumn{3}{c}{\textbf{Cross-pack experiment}} \\
\cmidrule(lr){1-3}
\textbf{ID} & \textbf{Model} & \textbf{Precision} \\
\midrule
Q4B  & Qwen3.5-4B                  & BF16 \\
Q9B  & Qwen3.5-9B                  & BF16 \\
Q27B & Qwen3.5-27B                 & BF16 \\
Q35B & Qwen3.5-35B-A3B             & BF16 \\
G4B  & Gemma 3 4B IT               & BF16 \\
G12B & Gemma 3 12B IT              & BF16 \\
G27B & Gemma 3 27B IT              & BF16 \\
-  & Qwen3.5-35B-A3B-FP8         & FP8  \\
\bottomrule
\end{tabular}
\end{minipage}

\end{table}

\subsection{Generation Parameters}

All local models were run with a shared sampling configuration
(Table~\ref{tab:gen-params}). Extended thinking was disabled for all runs.
The GPT-5 API endpoint was called with its default parameters.

\begin{table}[ht]
\centering
\caption{Sampling parameters applied to all local models.}
\label{tab:gen-params}
\begin{tabular}{ll}
\toprule
\textbf{Parameter} & \textbf{Value} \\
\midrule
Temperature        & 0.7 \\
Top-$p$            & 0.8 \\
Top-$k$            & 20 \\
Min-$p$            & 0.0 \\
Presence penalty   & 1.5 \\
Repeat penalty     & 1.0 \\
Chain-of-Thought  & disabled \\
\bottomrule
\end{tabular}
\end{table}

\subsection{Software}

Table~\ref{tab:software} records the exact software versions and commit hashes
used to generate all reported results.
The SimpleAudit \texttt{benchmarking} branch extends the main library with
data-capture and statistics; it does not alter the auditing or scoring logic.
Note that the Petri repository was relocated during development of this paper;
the canonical URL at time of writing is given in the table.

\begin{table}[ht]
\centering
\caption{Software used to generate results. Commit hashes pin the exact
revision.}
\label{tab:software}
\small
\resizebox{\textwidth}{!}{%
\begin{tabular}{llll}
\toprule
\textbf{Component} & \textbf{Commit} & \textbf{Branch} & \textbf{Repository} \\
\midrule

  SimpleAudit &  \texttt{904d10a} & \texttt{benchmarking} &
  \url{https://github.com/finnschwall/simpleaudit/tree/benchmarking} \\
SimpleAudit (upstream) & N/A & \texttt{main} &
  \url{https://github.com/kelkalot/simpleaudit} \\

Petri & \texttt{7fef276} & \texttt{petri-v2} &
  \url{https://github.com/meridianlabs-ai/inspect_petri} \\
llama.cpp & \texttt{d6f3030} & \texttt{main} &
  \url{https://github.com/ggml-org/llama.cpp} \\
\bottomrule
\end{tabular}
}
\end{table}

\subsection{Hardware}

Inference hardware was not held fixed across experiments. Because all local
model outputs are logged verbatim by SimpleAudit and archived in the analysis
repository, numerical results are fully reproducible from the stored records
without re-running inference.



\clearpage

\end{document}